\RequirePackage{fix-cm}

\RequirePackage{amsmath}  
\documentclass[twocolumn]{svjour3}          
\smartqed  
\usepackage{graphicx}
\usepackage{hyperref} 
\usepackage[dvipsnames,table]{xcolor}
\usepackage{mathtools}
\usepackage{amsmath} 
\usepackage{algorithm}
\usepackage{amssymb}  
\usepackage{caption} 
\usepackage{algpseudocode}
\usepackage{nccmath} 
\usepackage{algorithm}
\usepackage[normalem]{ulem}
\usepackage{rotating} 
\newcommand{\doubleBlind}[1]{} 

\newcommand{\lt}{{\tt True}}
\newcommand{\lf}{{\tt False}}

\newfloat{spec}{thb}{lop} 
\floatname{spec}{Specification}
\makeatletter
\newcommand{\leqnomode}{\tagsleft@true}
\newcommand{\reqnomode}{\tagsleft@false}
\makeatother

\begin{document}





\title{Accomplishing High-Level Tasks with Modular Robots
}


\author{Gangyuan Jing \and Tarik Tosun \and Mark Yim \and Hadas Kress-Gazit}


\institute{G. Jing \at
           Cornell University \\
           \email{gj56@cornell.edu}           
           \and
           T. Tosun \at
           University of Pennsylvania\\
           \email{tarikt@seas.upenn.edu}
           \and
           M. Yim \at
           University of Pennsylvania\\
           \email{yim@seas.upenn.edu}
           \and
           H. Kress-Gazit \at
           Cornell University  \\
           \email{hadaskg@cornell.edu}
}
\date{Received: date / Accepted: date}

\maketitle


\begin{abstract}
The advantage of modular self-reconfigurable robot systems is their flexibility, but this advantage can only be realized if appropriate configurations (shapes) and behaviors (controlling programs) can be selected for a given task.  In this paper, we present an integrated system for addressing high-level tasks with modular robots, and demonstrate that it is capable of accomplishing challenging, multi-part tasks in hardware experiments. The system consists of four tightly integrated components: (1) A high-level mission planner, (2) A large design library spanning a wide set of functionality,  (3) A design and simulation tool for populating the library with new configurations and behaviors, and (4) modular robot hardware. This paper build on earlier work by the authors \cite{jing2016end}, extending the original system to include environmentally adaptive parametric behaviors, which integrate motion planners and feedback controllers with the system.
%
%
\end{abstract}

%
\setlength\tabcolsep{0pt}
\begin{figure}
\begin{center}
\begin{tabular}{c c c}
\includegraphics[width=0.32\columnwidth]{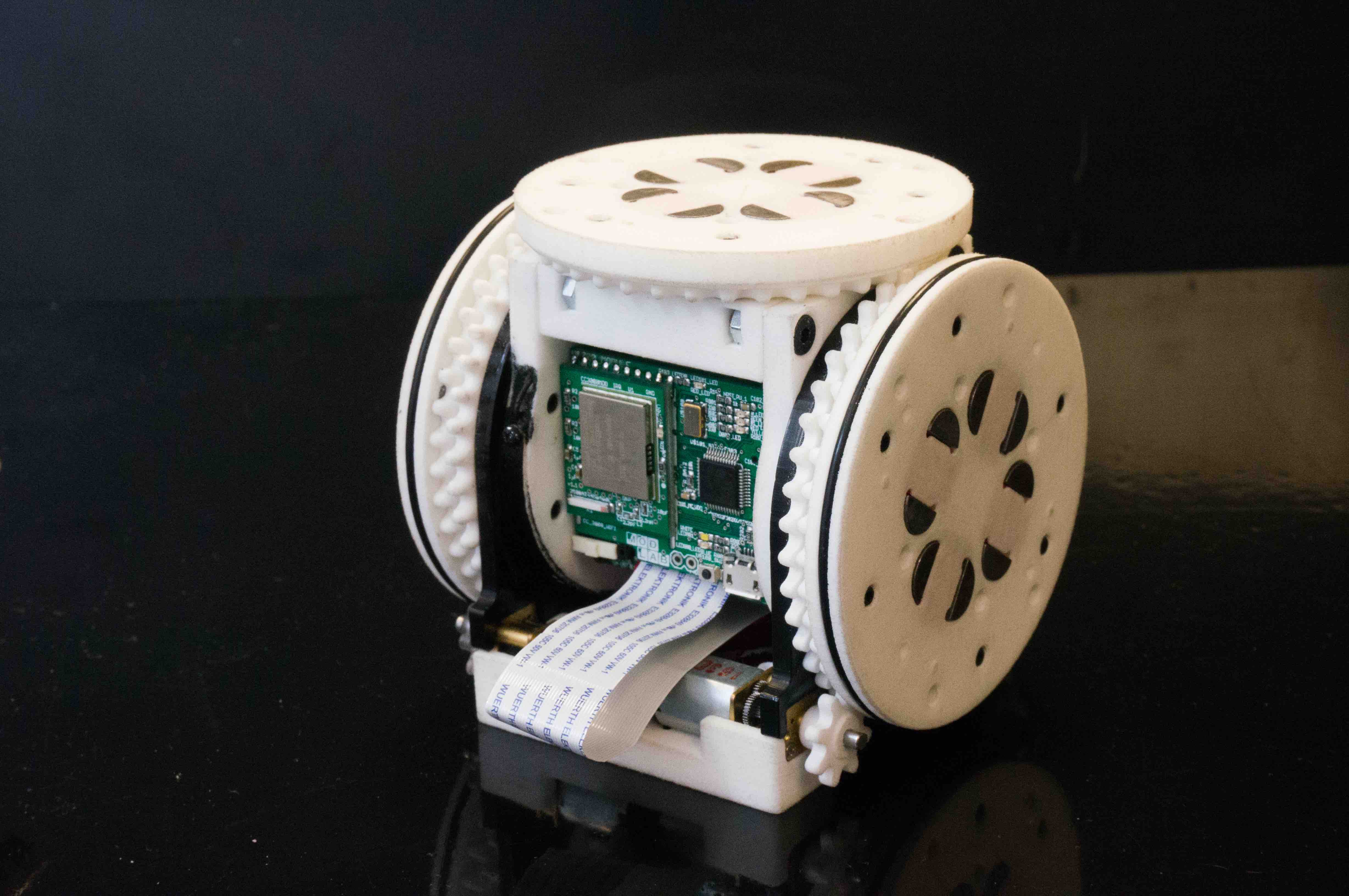}&
\includegraphics[width=0.32\columnwidth]{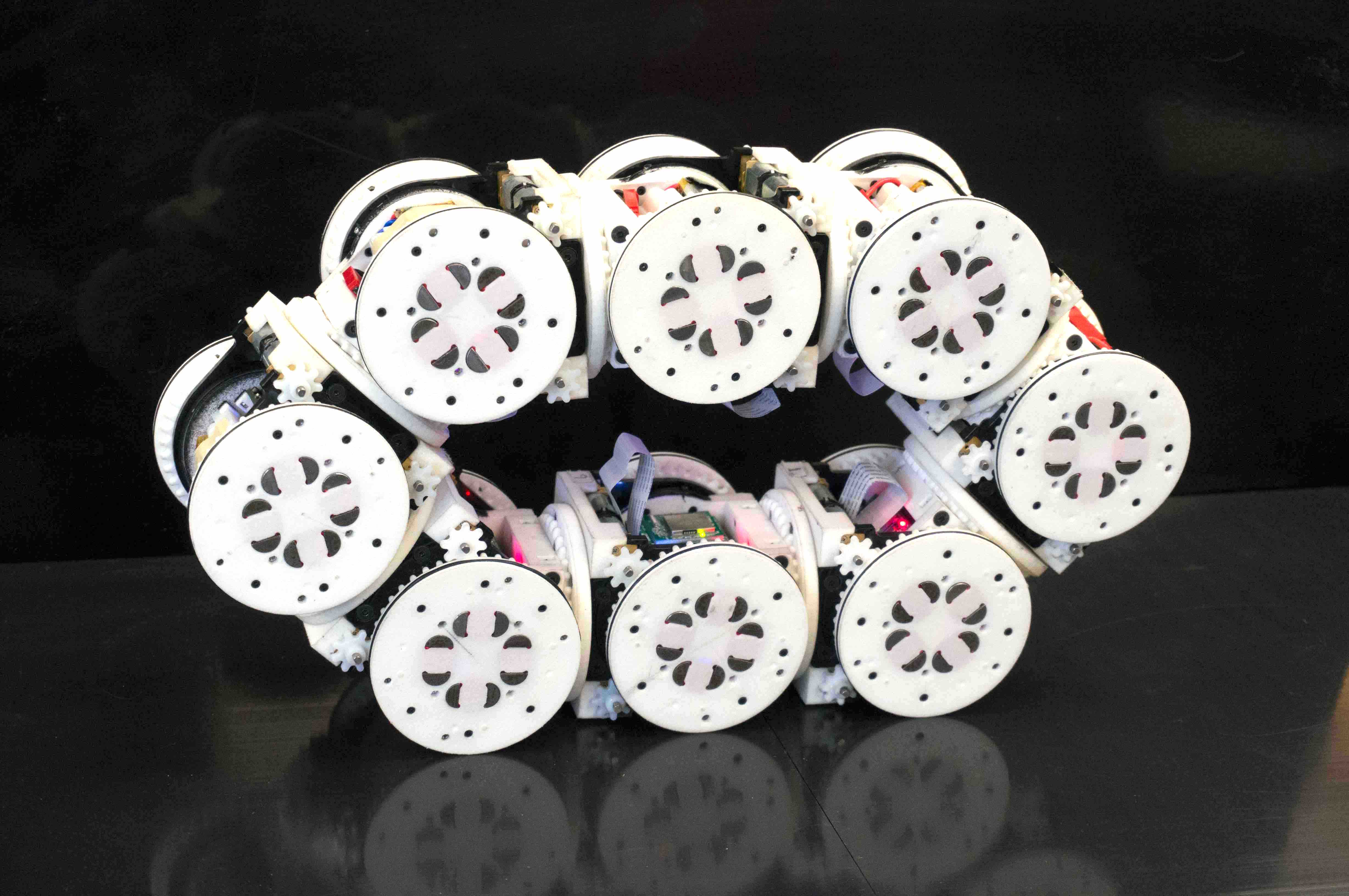}               
&
\includegraphics[width=0.32\columnwidth]{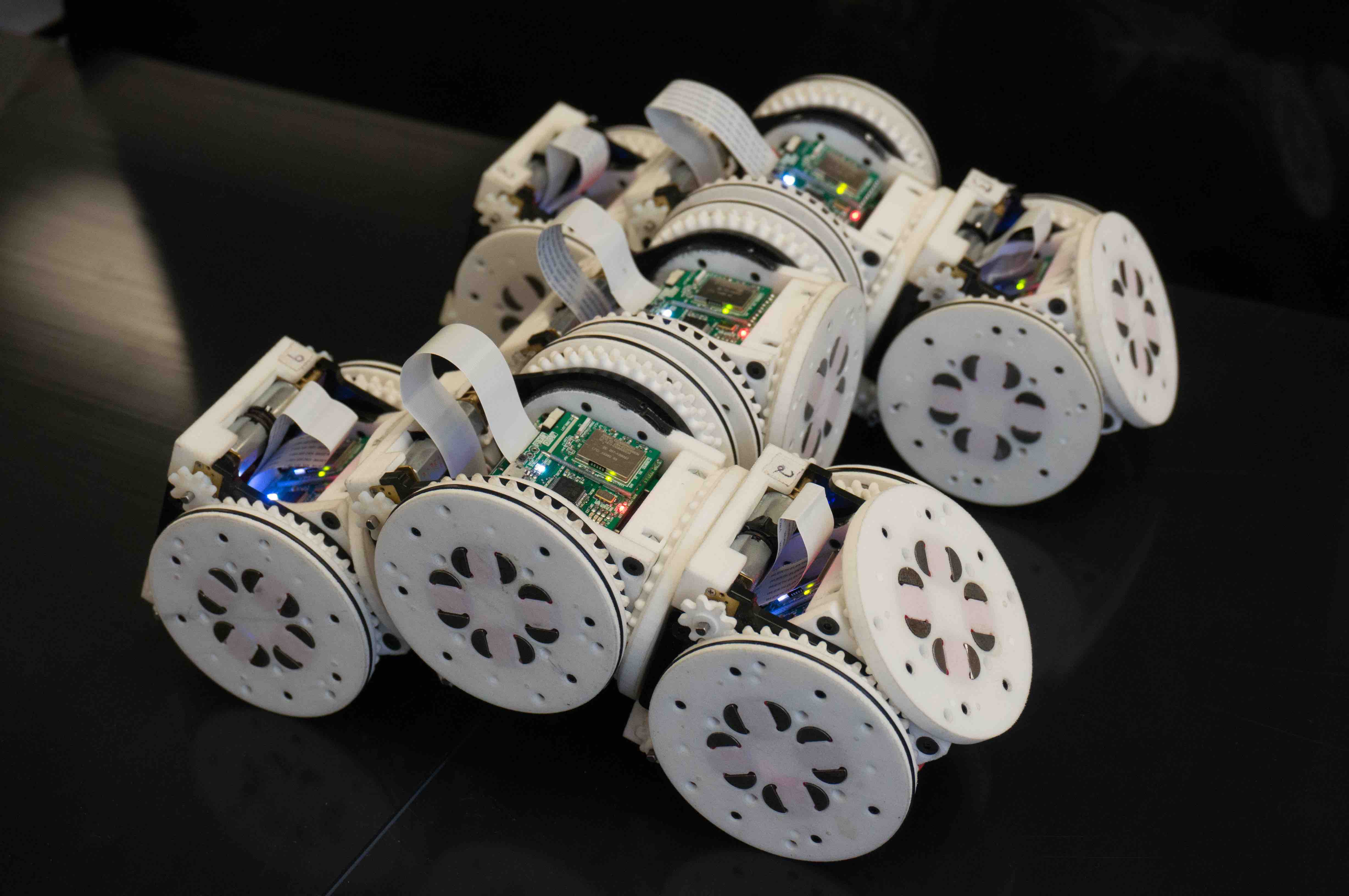} \\ 
\includegraphics[width=0.32\columnwidth]{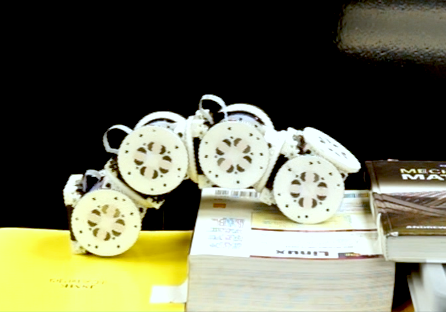} & 
\includegraphics[width=0.32\columnwidth]{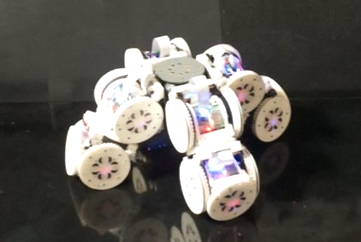} &
\includegraphics[width=0.32\columnwidth]{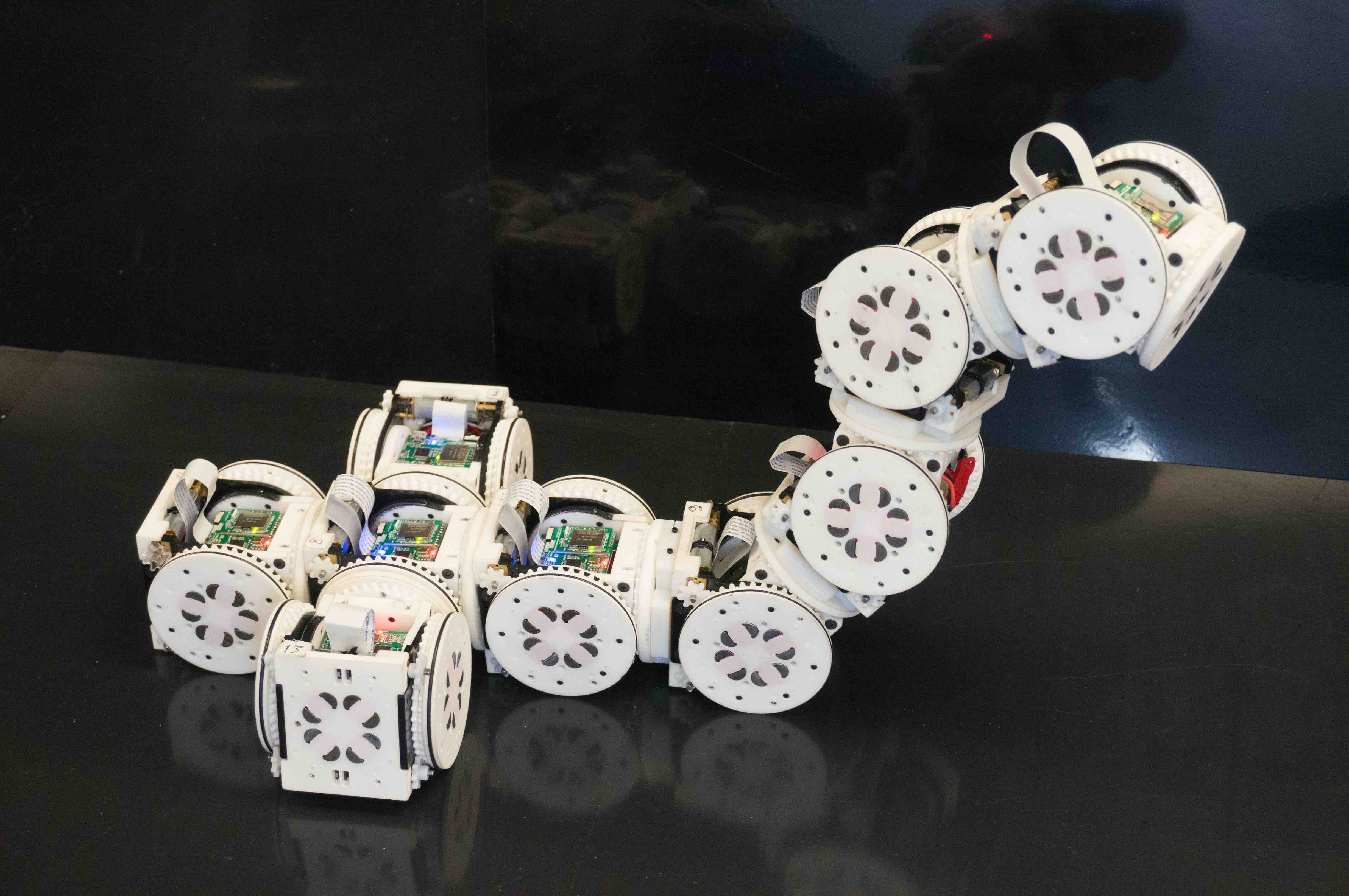} 
\end{tabular}
\caption{Six configurations from the design library}
\label{fig:designs}
\end{center}    
\end{figure}
\section{Introduction} \label{sec:introduction}
Modular self-reconfigurable robots (MSRR) are systems composed of repeated robot elements (called \emph{modules}) that have the ability to connect together to form larger robotic structures. These systems distinguish themselves from traditional robot systems through their ability to \emph{self-reconfigure}: changing the connective structure of the modules to assume different shapes that have different capabilities. Over the last three decades, many kinds of modular reconfigurable robots have been built \cite{atron}, \cite{mtranIII}, \cite{fukuda1990cellular},\cite{lipson2000towards}, and many different approaches have been introduced for controlling and programming them \cite{salemi2001hormone}, \cite{stoy2002using}, \cite{zhang2003phase}.

Robotics research is increasingly focused on deploying robots in real-world applications such as search and rescue.  Operating in these scenarios entails handling an enormous amount of variability in task requirements and environment conditions.  One approach to this problem is to build complex monolithic systems, such as large humanoids \cite{kuindersma2016optimization}.  These systems can perform a wide range of actions, but are extremely complex.  In a sense, their broad range of capability comes at the cost of having to solve each individual problem in a complicated way. For example, to pick up and move an object, a humanoid must balance on two legs while using a high degree of freedom (DOF) arm to manipulate the object. In contrast, a robot
that was purpose-built for that task could accomplish it with far fewer DOF, and would require less complicated control algorithms.

The strength of MSRR systems lies in their flexibility. In principal, self-reconfiguration will allow modular robots to transform into designs specifically tailored to the needs of each new task they encounter, allowing them to elegantly address a wide variety of tasks by reconfiguring into a wide variety of  solutions.  However, this strategy poses an obvious challenge: given a task, is it possible to select an appropriate configuration (robot shape) and behavior (controlling program) to address it?  This is an important unsolved problem in the field, and remains a significant barrier to the use of modular robots to solve real-world problems \cite{yim2007modular}.

In this paper, we present a system capable of selecting appropriate modular robot configurations and behaviors to solve complex high-level tasks. Our system is library-driven: rather than attempting to generate new designs from scratch, users specify task requirements and a high-level controller retrieves  designs satisfying the requirements from a library of existing designs.  In addition to library management, the system integrates tools for low-level design creation, high-level mission planning, and physical modular robot hardware.

We leverage ideas from recent work on automatic controller synthesis with correctness guarantees from high-level task specification \cite{belta07,bhatia2010sampling,Kloetzer08,KGFP_TRO09,Raman15,Wongpiromsarn10}.  These methods have proven effective for addressing high-level tasks with traditional robots, allowing users to specify task requirements at a high level using formal languages and then automatically synthesizing low-level robot controllers with performance guarantees. Applying these methods in the context of modular robotics introduces an additional layer of complexity due to the fact that the morphology of the robot is not fixed.

This paper builds upon our earlier work, presented in \cite{jing2016end}. Specifically, we expand our system by introducing environmentally-adaptive parametric behaviors (EAP behaviors) that can leverage sophisticated motion planners and feedback controllers to continuously respond to environment conditions.

Through hardware experiments, we demonstrate that our system is capable of addressing challenging multi-part tasks. This paper presents the details of the system, discusses its strengths and weaknesses, and provides a roadmap forward to apply a similar system in a  real-world setting. 
We hope others will be able to adopt our framework and utilize it to bring modular robots into real-world applications.
%
\subsection{System Overview} \label{sec:system-overview}
%
\begin{figure}[h]   
    \begin{center}
    \includegraphics[width=\columnwidth]{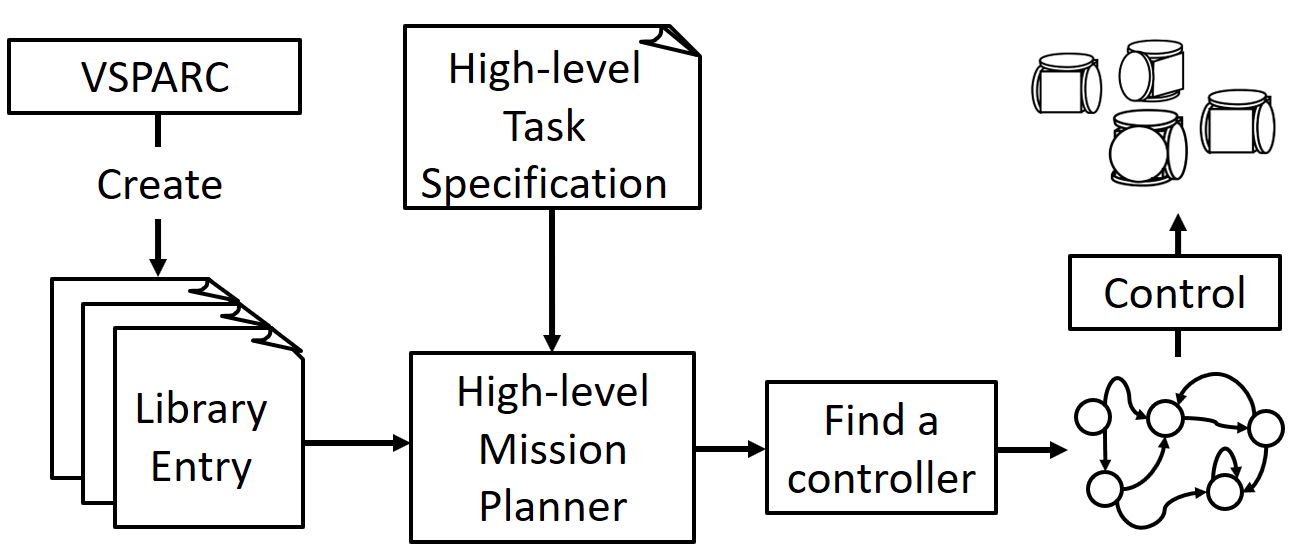}
    \end{center}
    \caption{System flowchart}
    \label{fig:flowchart}
\end{figure}

Here, we provide a brief overview of the entire system. Figure~\ref{fig:flowchart} provides a visual companion to this section.

The system is built around a design library that spans a wide range of useful functionality.  Library entries are configurations and behaviors for the SMORES-EP modular robot, which are designed in a physics-based simulator and design tool called VSPARC which we created for this purpose.  Users build, program, and test modular robot designs through a graphical user interface, and can save their designs to a web server, allowing them to be shared with others.  Any configuration or behavior created in the simulator can be directly ported to the hardware modular robot system, SMORES-EP.

Our system allows users to solve high-level tasks with modular robots.  Tasks are specified in a mission planning tool using Structured English~\cite{Finucane_IROS10}, a high-level language.  Users do not specify which configurations and behaviors should be used to complete the task, but rather describe the required functionality. For example, the user might request that the robot perform a \textbf{drive} action in a \textbf{tunnel} environment labeled with the property $\texttt{max\_height}=3$. 

To develop a solution to the task, the high-level mission planner fulfills each of the specified functionality by automatically selecting robot configurations and behaviors from the design library, generating a  controller in the form of a finite-state automaton.  In the above example, the system could select any configuration that is capable of executing a \textbf{drive} behavior while maintaining a maximum height of 3 modules or less. In a sense, the high-level planner treats the entire modular robot system as a single robot with a set of capabilities defined by the library.
The mission planner can then execute the controller to complete the task, directly commanding hardware SMORES-EP robots based on environment information from sensors.
\subsection{Contributions}
This paper presents an integrated system capable of addressing high-level tasks with modular robots.  The tasks it addresses are \emph{reactive}: they require decision-making about what action to perform based on the sensed environment; \emph{complex}: they include multiple sub-tasks with potentially very different requirements; and \emph{high-level}: the task specification encodes the desired outcomes, and the system intelligently synthesizes a solution that results in those outcomes using the available configurations and behaviors from the library.  

This system is one of the first to address these kinds of tasks with modular self-reconfigurable robots, which introduce an additional layer of complexity because they can assume many configurations.  This represents a significant contribution to the field, because such systems will be necessary for modular robots to operate in realistic task scenarios. By providing this framework and demonstrating its success in the lab, we hope to lay the foundation for future modular robot systems to address tasks in the real world.

The system  includes four tightly integrated components: (1) A high-level mission planner, (2) A large design library spanning a wide set of functionality, (3) A design and simulation tool for populating the library with new configurations and behaviors, and (4) modular robot hardware.  Several of the subcomponents represent research contributions. Our novel design tool (VSPARC) represents a novel contribution, as does our library of 52 configurations and 97 behaviors.  We also introduce a minor theoretical contribution by checking the feasibility of robot behaviors prior to controller synthesis.

This paper builds on earlier work by the authors, presented in \cite{jing2016end}.
 In Section~\ref{sec:parametric}, we present environmentally adaptive parametric
behaviors, which are a major novel extension to the system presented in \cite{jing2016end}.
 These behaviors  allow sophisticated
closed-loop behaviors to be developed, integrating motion planners and feedback control.
%
\section{Related Work} \label{sec:related-work}
Modular self-reconfigurable robots (MSRR) systems have the potential for great advantage over traditional robot systems in scenarios where flexibility is required: for example, search-and-rescue scenarios where the environment and task requirements may not be well-known \emph{a priori}.  Much of the existing research in MSRR systems has focused on establishing the fundamental capabilities that differentiate these systems from traditional robots.  Notably, MSRR systems have demonstrated the ability to form a wide variety of physical morphologies capable of diverse modes of locomotion, suitable to a range of different terrains \cite{Yim1994}.  The ability to autonomously reconfigure has been demonstrated \cite{yim2007towards}, and a number of reconfiguration planning algorithms have been developed \cite{sung2015reconfiguration}.

Similarly, a great deal of work has been done to develop behaviors for MSRR.  Many efforts focus on distributed control strategies, taking advantage of the distributed nature of MSRR hardware \cite{walter2002choosing}.  Distributed strategies include central pattern generators \cite{sproewitz2008learning} and hormone-based control \cite{salemi2001hormone}.  Genetic algorithms have been used to automatically generate both modular robot designs and behaviors \cite{hornby2003generative}.

It is clear that MSRR systems have demonstrated the ability to accomplish low-level tasks such as reconfiguration, locomotion, and manipulation.  However, to truly live up to their promise of flexibility in real-world applications, systems must be developed that leverage these low-level capabilities to address complex, high-level, multi-part tasks.

While there is a robust body of research into addressing high-level tasks with traditional robots, 
little work has been done in this area with modular robots. High-level control of modular robots poses a unique challenge, because solving tasks involves selecting not only appropriate behaviors, but also appropriate configurations.  This makes it all the more important to develop automated systems that can synthesize task-appropriate modular robot configurations and behaviors from high-level specifications.

In \cite{castro2011high}, Castro \emph{et al.} introduce a high-level control framework for the CKBot modular robot.  This framework lays the theoretical foundations for our high-level mission planner, one of the four major components of our system.  We expand the framework into a larger system capable of addressing significantly more sophisticated tasks. In addition to the mission planner, we provide design and simulation tools for creating and testing modular robot configurations and behaviors, and a large library (52 designs, 97 behaviors, 19 properties) with designs capable of addressing a wide range of tasks.  We expand the theoretical formalism introduced by Castro to include both behavior and environment properties, increasing the expressiveness of task specification, and introduce a performance improvement by grounding abstract action specifications in concrete configurations and behaviors prior to automata synthesis.

Tosun \emph{et al.} \cite{tosun2015computer} introduce a system that allows users to rapidly synthesize modular robot designs and behaviors by composition.  The system includes a physics-based simulator and a hierarchically organized library of configurations and associated behaviors.  The goal of this work is to aid in the selection of modular robot configurations and behaviors appropriate to complex tasks,  but it takes a very different approach than our automated system, instead providing tools for users to manually create new designs by combining library entries using series and parallel composition operations. 

Outside the realm of modular robotics, systems have been developed that can synthesize rapidly manufacturable robot designs from high-level user specifications \cite{mehta2014cogeneration},\cite{mehta2014design}, \cite{schulz2015interactive}.  This work is similar to ours in the sense that high-level specifications from the user are interpreted to synthesize robot designs and behaviors from elements in a design library. The goal of these systems is to allow novice users to rapidly design and build functioning robots at low cost, using fabrication techniques such as 3D printing \cite{mehta2014cogeneration},\cite{mehta2014design} and origami folding \cite{schulz2015interactive}.  Consequently, the scope of the tasks they address is very different from ours.  In these systems, library entries are electromechanical components such as motors, motor drivers, and microcontrollers, and a high level task might be ``Create a robot that can walk and turn.''  In contrast, library entries in our system are whole robots with associated behaviors, and we address complex, multi-part tasks such as ``Climb on top of the table, and move any debris you find into the trash bin.''
%
\section{Background} \label{sec:background}
In this section, we define modular robot systems and provide background on controller synthesis techniques.

\subsection{Modular Robot Systems}
\begin{definition}[Module]
A module is the fundamental unit of a modular robot system.
Each module is a small robot that can receive and respond to commands, move, and connect to other modules. In this work, we consider only homogeneous modular robot systems, meaning that all modules in the system are identical.

We define a module as $m=(J, A)$. 
$J=\{J_{1}, \dots, J_{d}\}$ is the set of joints of the module with $d$ degrees of freedom.
$A=\{A_1, \dots, A_k\}$ is the set of {\it attachment points} where the module can connect to other similar modules.
Each attachment point can only connect to one other module at a time.
We denote the attachment point $A_i$ of module $m$ as $m.A_i$. 
\end{definition}

\begin{definition}[Configuration]
A configuration is a connected set of modules that acts together as a single robot.
 The smallest configuration is a single module.
A configuration is denoted as $\mathcal{C}=(M, E)$, where $M=\{m_1,\dots,m_q\}$
is the set of connected modules that form the configuration and
$E$ is the set of connections between modules, represented as pairs of attachment points $(m_{i}.A_{a_1},m_{j}.A_{a_2})\in
E$, where $m_{i},m_{j}\in M$, and $m_{i}\neq m_{j}$.
\end{definition}

\begin{definition}[Joint Command]
Joint commands are used to control the joints of the modules.
A command to a joint $J_i$ is defined as $u_{J_i}=(\alpha, V, t)$, where $\alpha\in\{\text{Position, Velocity}\}$ is the type of command, $V\in \mathbb{R}$ is the value of the command, and $t\in \mathbb{R}$ is the time duration of the command.
For example, $u_{J_i} = (\text{Position},\frac{\pi}{2},2)$ commands joint $J_i$ to hold the angle $\theta=\frac{\pi}{2}~rad$ for $2$ seconds. 
Similarly, $u_{J_i} = (\text{Velocity},\pi,3)$ will drive joint $J_i$ with angular velocity of $\dot\theta=\pi~\frac{rad}{sec}$ for $3$ seconds.
We assume there are low-level controllers (\textit{e.g.} PID controllers) that
can drive the corresponding joint to satisfy the command $u_{J_i}$.
\end{definition}

\begin{definition}[Behavior]
For a configuration $C$, we define a behavior $B_C= \left\{ b_1,\dots,b_n \right\}$ as a sequence of behavior states.
Each behavior state is defined as $b_i=(U, T)$, where $U$ is the set of joint commands
for all joints of all modules in the configuration. The time duration $T$ of each behavior state is equal to longest duration of its joint commands $U$, ensuring that behavior execution will move on to the next state only once all joint commands in the current behavior state have completed.
\end{definition}

\subsection{Controller Synthesis}\label{sec:backgroundsynthesis}
In this work, we utilize existing work on controller synthesis \cite{Finucane_IROS10,KGFP_TRO09} to generate high-level controllers for modular robot systems.
The process of controller synthesis consists of three main steps:
(1) representing the robot and the environment using a discrete abstraction,
(2) expressing desired robot tasks with a formal specification language,
(3) searching for a control strategy that satisfies the given task specification, or determining that such a strategy does not exist. 

\paragraph{\bf Robot and Environment Abstraction:}
To represent the continuous environment state and robot actions as discrete models,
we abstract the environment events and robot capabilities into sets of boolean variables. 
The value of each variable represents the sensed environment state or the current robot actions.
For example, the environment variable {\bf Cup} is $\lt$ if and only if the robot is currently sensing a cup with its camera.
Similarly, the robot variable {\bf Push} is $\lt$\ if and only if the robot currently performing a pushing action.

\paragraph{\bf Robot Task Specification:}
A wide range of robot tasks can be defined using a formal language called Linear Temporal Logic (LTL).
In \cite{Finucane_IROS10}, authors introduce a tool called LTLMoP that allows users who are unfamiliar with LTL to specify robot tasks in a formal language called Structured English, which is closer to natural language.  LTLMoP then automatically translates Structured English specifications into LTL formulas.
The following is an example of a robot task specification written in the Structured English:
\begin{itemize}
\item visit {\bf Classroom}
\item if the robot senses {\bf Student} then do {\bf Greet}
\item do {\bf Pickup} if and only if the robot senses {\bf Trash}
\end{itemize}

In these examples, {\bf Student} and {\bf Trash} are environment variables,  while {\bf Classroom}, {\bf Greet}, {\bf Pickup} are robot action variables. 
To connect the high-level specification with physical robot systems, users  provide mappings from robot action variables to low-level robot controllers, and from environment variables to sensors.

\paragraph{\bf Controller Synthesis and Execution:}
Authors of \cite{KGFP_TRO09} introduce a framework to automatically generate a high-level robot controller to satisfy a task specification, or decide such controller does not exist.
The synthesized controller is a finite-state automaton, and specifies robot actions that satisfy the task.
Each state in the controller is labeled with robot variables, and each transition is labeled with environment variables.

To accomplish a tasks, the synthesized controller is implemented continuously by mapping each robot variable to a low-level robot controller, and mapping each environment variable to a robot sensing function.
With these mappings, the robot is able to detect the environment and perform desired actions to satisfy the task specification.

%
\section{System} \label{sec:system}
%
%
\subsection{Modular Robot Hardware - SMORES-EP Robot}
\label{sec:hardware}
Our system is built around the SMORES-EP modular robot, but could be extended to other modular robot hardware systems. In this section we provide an overview of the capabilities of SMORES-EP. 

\begin{figure}[h]   
    \begin{center}
    \includegraphics[width=\columnwidth]{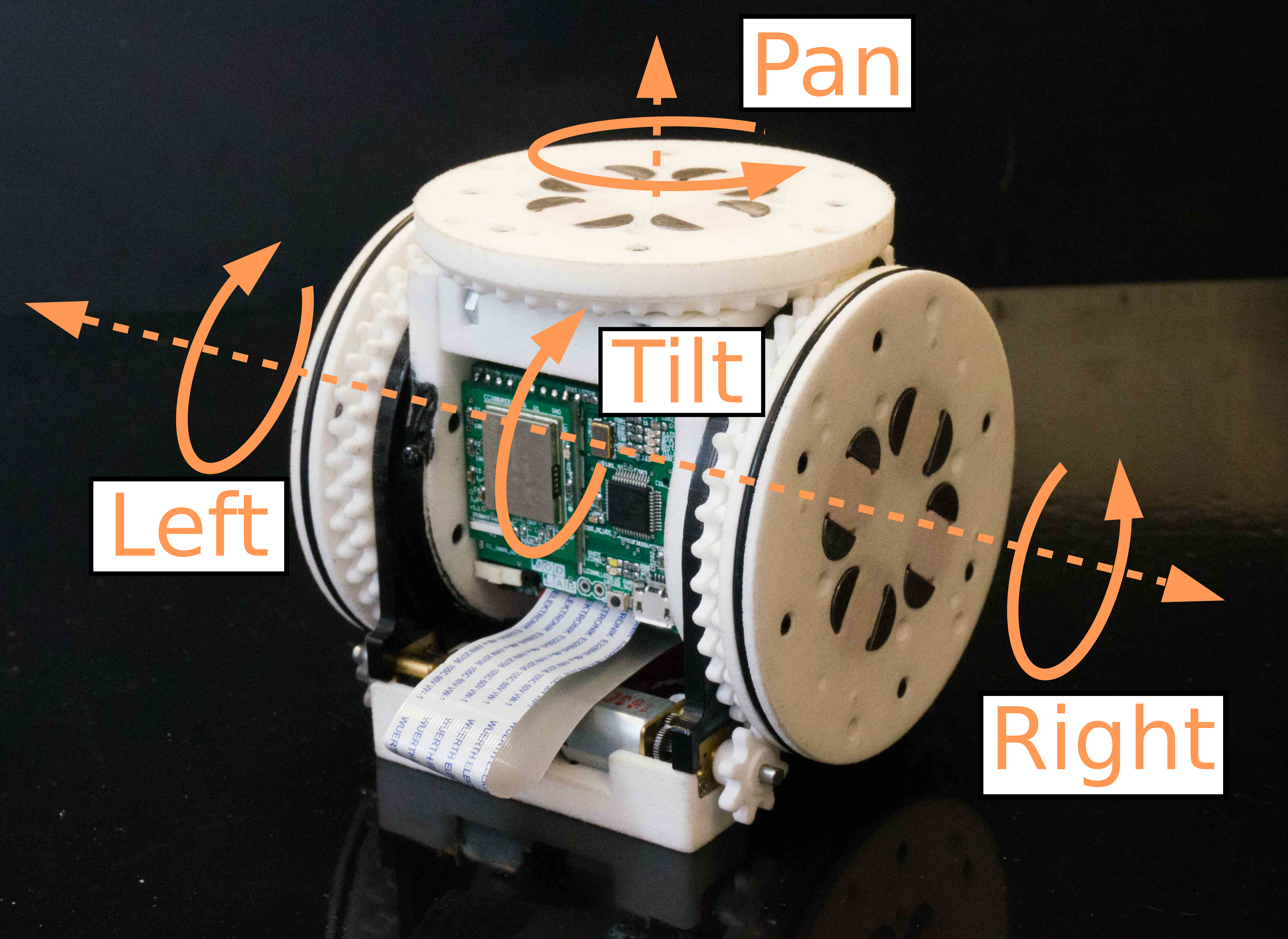}
    \end{center}
    \caption{SMORES-EP module}
    \label{fig:smores-module}
\end{figure}

Each SMORES-EP module (Figure~\ref{fig:smores-module}) is the size of an $80mm$ cube.  Four faces of the cube have magnetic connectors known as \emph{EP-Faces} that allow them to connect to other modules.  The EP-Face connector is an array of four electro-permanent magnets (EP magnets) embedded in a planar face, and provides fast, strong, energy-efficient connection between modules. Each EP magnet consist of an electromagnet coil wrapped around a core of two permanent magnet rods.  Short pulses of current through the coil generate a magnetic field that re-polarizes one of the magnets in the core, allowing the external force to be turned on (magnetically attractive) or off (neutral).  Once polarized, the magnets will maintain either state indefinitely, so a pair of connected EP-Face sustains a connection strength of $88.4N$ without consuming any power.  Each face requires $80ms$ and $2.5J$ of energy to switch states. The magnets are arrayed in a ring with south poles counterclockwise of north, making the connector hermaphroditic (any two faces can connect) and able to connect at $90^\circ$ increments \cite{Tosun2016}.

The modules are kinematically identical to their predecessor, the SMORES robot \cite{davey2012emulating}, and have four actuated joints. The \emph{left} and \emph{right} faces of the module are able to rotate continuously at a maximum rate of $90^\circ$ per second and can be used as wheels, allowing individual modules to move by differential drive.  These faces have thin rubber tires to enable driving on a variety of surfaces. The circular top face is also able to rotate continuously at a maximum rate of $30^\circ$ per second, and is referred to as the \emph{pan} joint. A central bending joint (referred to as the \emph{tilt} joint) has a $180^\circ$ range of motion, allowing the top face to bend forward or backward until it is perpendicular to the bottom face.  Each of the joints is equipped with a custom potentiometer for position sensing \cite{Tosun2017paintpots}, and modules perform feedback control for all joints at a rate of 20Hz.

The motions that a SMORES-EP cluster can perform are limited by the strength of the motors and connectors.  A pair of connected EP-Faces can sustain a maximum bending load of $1.8Nm$, equivalent to supporting 3.1 modules in cantilevered horizontally against gravity \cite{Tosun2016}.  If necessary, this limitation can be alleviated by installing connector plates that screw into the faces of two modules to rigidly connect them.  In this case, four modules can be supported in cantilever before the motor torque limits are exceeded.  When two modules are attached using connector plates, they lose the ability to disconnect and self-reconfigure.

The SMORES-EP system also includes passive cubes that can act as lightweight structural elements in SMORES-EP robot configurations.  These plastic cubes have the same $80mm$ form factor as modules, and have an array of 8 permanent magnets on each face, allowing them to make a strong connection to modules. 

Each module has its own $600mAh$ LiPo battery, microcontroller (STM32F303), and 802.11b WiFi module (TI CC3000), allowing it to move and operate independently or as part of a cluster.  Battery life is typically about one hour, assuming typical usage of the motors and and wireless communication.  The EP-magnets require very little energy, and usage does not significantly affect battery life.

In this work, a cluster of modules is controlled by a central computer running a Python program that sends wireless messages (UDP packets) to control the movement and magnets of each module.  Wireless networking is provided by a standard off-the-shelf router, with a range of about 100 feet.  Because individual SMORES-EP modules do not have any way of sensing their environment, localization is provided by AprilTag markers \cite{olson2011apriltag} mounted to modules and objects of interest, tracked by an overhead camera.  The AprilTag tracker, high-level planner, and module control software run with a control loop time of about 4Hz on a laptop a 2.4GHz processor and 4GB of RAM.

%
\subsection{Design and Simulation Tool: VSPARC}
\begin{figure}[h]   
    \begin{center}
    \includegraphics[width=\columnwidth]{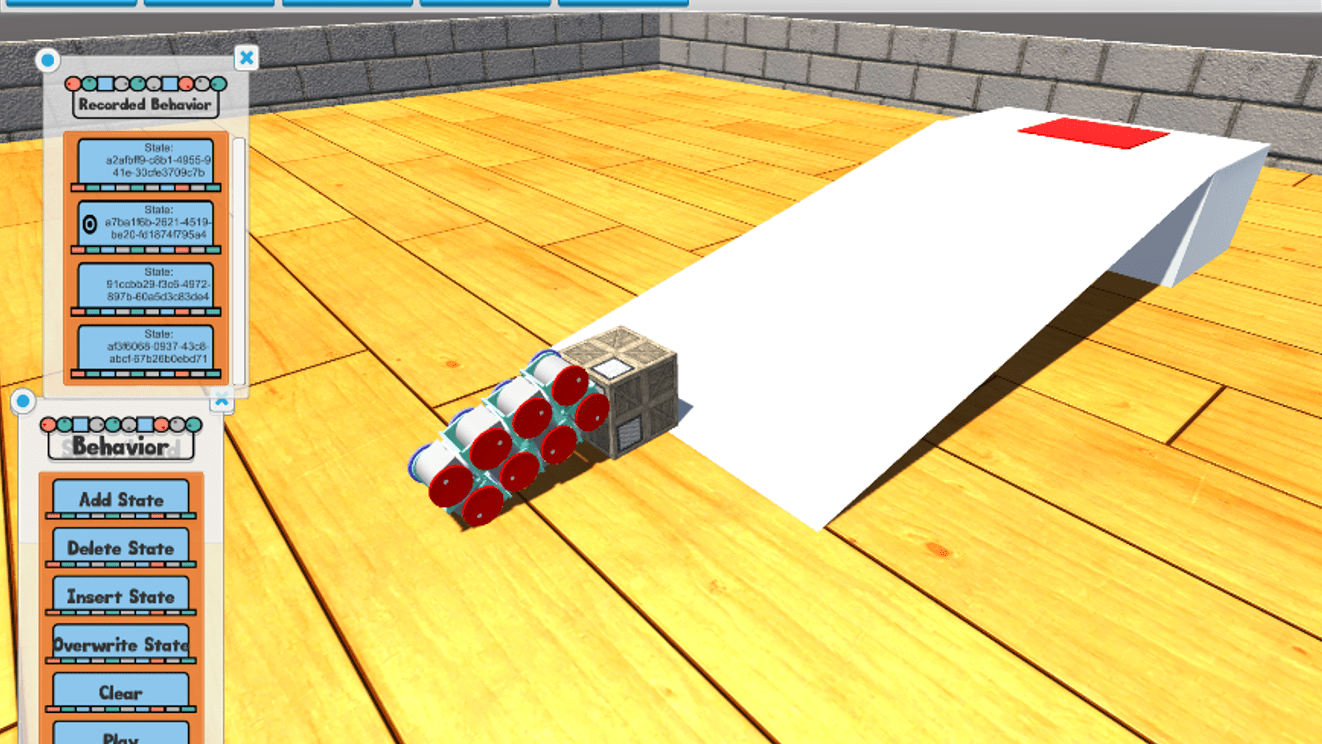}
    \end{center}
    \caption{VSPARC user interface}
    \label{fig:vsparcui}
\end{figure}

VSPARC, which stands for \textbf{V}erification, \textbf{S}imulation, \textbf{P}rogramming And \textbf{R}obot \textbf{C}onstruction, is our interactive design tool that allows users to design configurations and behaviors for SMORES-EP robots, and simulate them with a real-time physics engine.
As shown in Figure~\ref{fig:vsparcui}, the graphical user interface, powered by the Unity3D Engine \cite{Unity}, allows users with little background in robotics to design and test different robot configurations and behaviors.
The ability to control each joint of each module grants more experienced users the possibility to create complex designs.

%
VSPARC provides realistic physical modelling of \\SMORES-EP, taking into consideration factors such as the connector and actuator force limits.  This allows users to test and verify behaviors before running them with physical modules and receive early warning if, for example, their behavior would likely cause the connection between two modules to break.

\begin{figure}
    \includegraphics[width=3.3in]{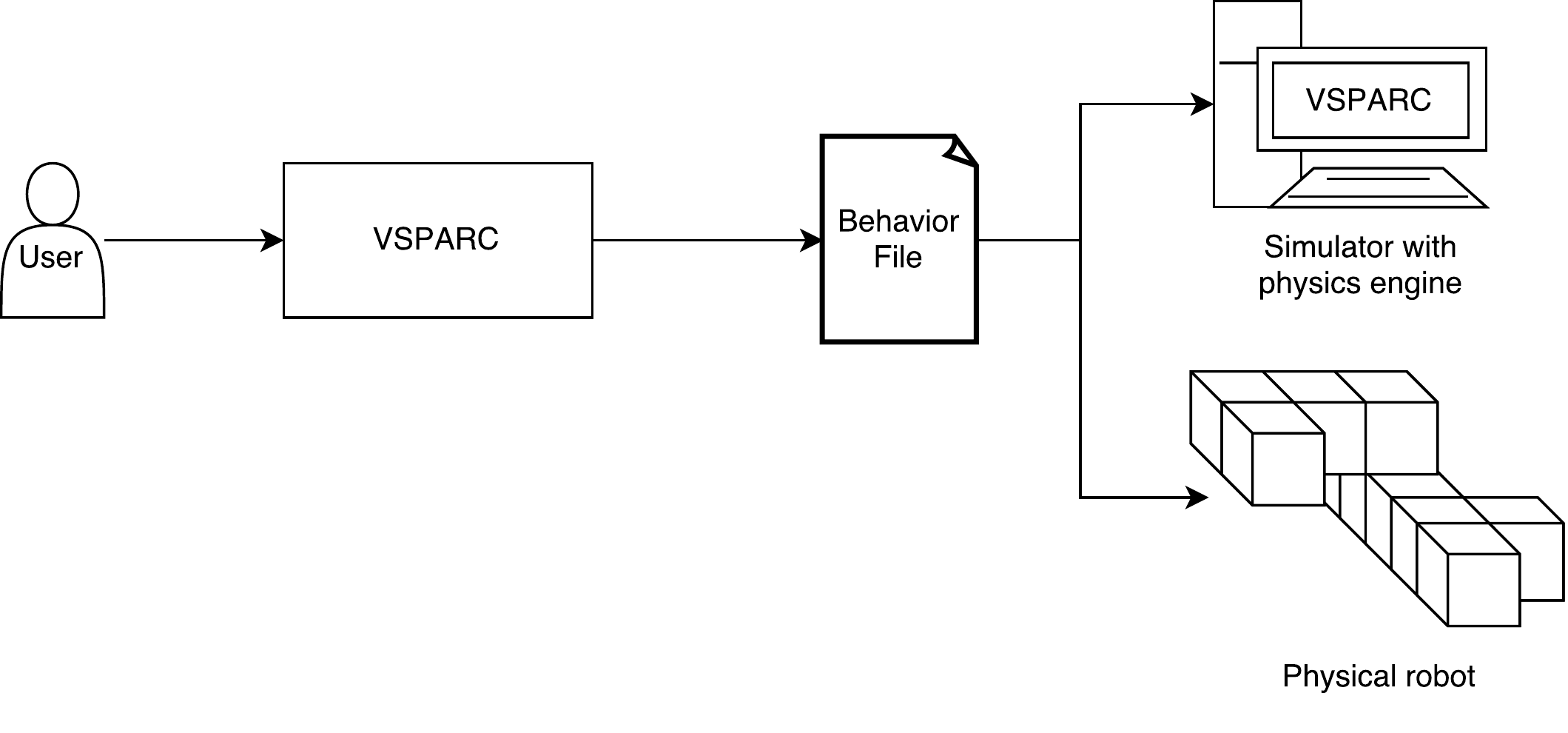}
    \caption{The same behavior file can be used by both the simulator and the physical robot.}
    \label{fig:vsparc1}
\end{figure}

VSPARC is available for free online at \url{www.vsparc.org}, and enables users to save and share their designs to a central server, allowing a large number of users to contribute to our design library.
VSPARC's main features are listed below:
\begin{itemize}
\item Design configurations with unlimited number of modules and visualize the design in a 3D environment.
\item Command positions or velocities for each joint of all modules.
\item Design behaviors for any configuration by creating a sequence of joint commands.
\item Simulate the performance of any behavior in a physics engine.
\item Create and share designs online. Test and improve other users' designs. 
\end{itemize}
As shown in Figure~\ref{fig:vsparc1}, behaviors designed in VSPARC can be exported as XML files and then run on SMORES-EP modules, providing seamless translation of behaviors from the simulator to physical robots.
%
\subsection{Design Library}
In this section, we introduce a library-driven framework to organize configurations and behaviors created in VSPARC. We introduce the notion of \textit{properties}, which specify the functionality and constraints of behaviors, and the \textit{robot design library}, which can be searched to find configurations and behaviors with desired properties.  

\begin{definition}[Property] \label{def:property}
Properties provide  high-level descriptions of the intended effects of a behavior, as well as the environment in which the behavior is appropriate.
We define a property as $p=(p_n, \Omega)$, where $p_n$ is the name of the property (i.e. a description title, in English)  and
$\Omega$ is the set of values of the property.
For example, a behavior with the  property $p=(\text{Action}, \{\text{Move, Push}\})$ can perform both {\it Move} and {\it Push} actions.
Properties are also used to describe the environmental conditions required for the behavior to run as expected.
For example, the property $p=(\text{ObjectWeight}, [2,5])$ indicates that the behavior can appropriately interact with an object if its weight is  between 2 and 5 module-weights.
In this case, the property is a quantitative description of the environment.
We say a property $p_1=(p_{n_1},\Omega_1)$ {\it satisfies} a property $p_2=(p_{n_2},\Omega_2)$
if and only if $p_{n_1}=p_{n_2}$ and $\Omega1 \subseteq \Omega2$.
\end{definition}

Properties connect tasks with behaviors that are appropriate to address them.
In Section~\ref{sec:matchprops}, we discuss how correct behaviors for a task can be automatically selected based on requirements over property values. Table~\ref{table:properties} lists some examples of environment and behavior properties that might be used for
common robot tasks. In the library, the unit for length is the side length of a single SMORES-EP module, and the unit for mass is the mass of a single SMORES-EP module.

\begin{center}
\captionof{table}{Examples of property names} 
\begin{tabular}{| c |c| c |} 
  \hline                        
  \textbf{Properties for} &&  \textbf{Properties for} \\
  \textbf{Robot Behavior} && \textbf{Environment} \\
  \hline 
  Speed && Box\_Mass \\
    \hline 
  Width && Stair\_Height \\
  \hline   
  Height && Ground\_Roughness \\
  \hline   
  Action && Tunnel\_Height \\
  \hline 
\end{tabular}\label{table:properties}
\end{center}

\begin{definition}[Robot Design Library]
 The design library is a collection
of modular robot configurations and behaviors labeled with environment and robot behavior
properties.
The library $\mathcal{L}$ consists of a set of library entries, $\mathcal{L}=\{l_1,l_2,\dots\}$.
Each library entry is defined as $l=(C, B_C, P_e, P_r)$, where $C$ is the configuration
and $B_C$ is a behavior associated with $C$.
$P_e$ and $P_r$ are sets of properties that describe the environment conditions and robot
behavior functionality, respectively.

As an example, the library entry:
\begin{eqnarray*}
&l = (C=\texttt{snake}, B_C=\texttt{climb}, P_e, P_r)\\
\mathrm{where:}~ &P_e = \{ ( \text{Ledge\_Height},[2,3]) \}\\
\mathrm{and:}~ &P_r =\{ ( \text{Action},[\text{Climb}]),(\text{Speed},[1]) \}
\end{eqnarray*}
represents a {\tt snake} shape configuration with  a {\tt climb} behavior that
can climb a ledge with a height of two to three module-lengths, with the speed of 1
module-length per second.
Moreover, we say a library entry $l$ {\it satisfies} a property $p$ if there exist a property $p'\in P_e\cup P_r$ such that $p'$ satisfies $p$.
\end{definition}

To populate the library with different configurations and behaviors designs, we made our design tool available online at \url{www.vsparc.org} and distributed the tool to undergraduate and graduate student volunteers, hosting three hackathons in which participants created designs for various robot tasks.
Currently, the library includes 52 configurations and 97 behaviors contributed by 20 volunteers. 
Since the full library is too large to list in this paper, we provide a representative sampling of configurations, behaviors, and properties in Table~\ref{tab:library-table}.
%
\begin{table*}
\center
        \includegraphics[height=\textheight]{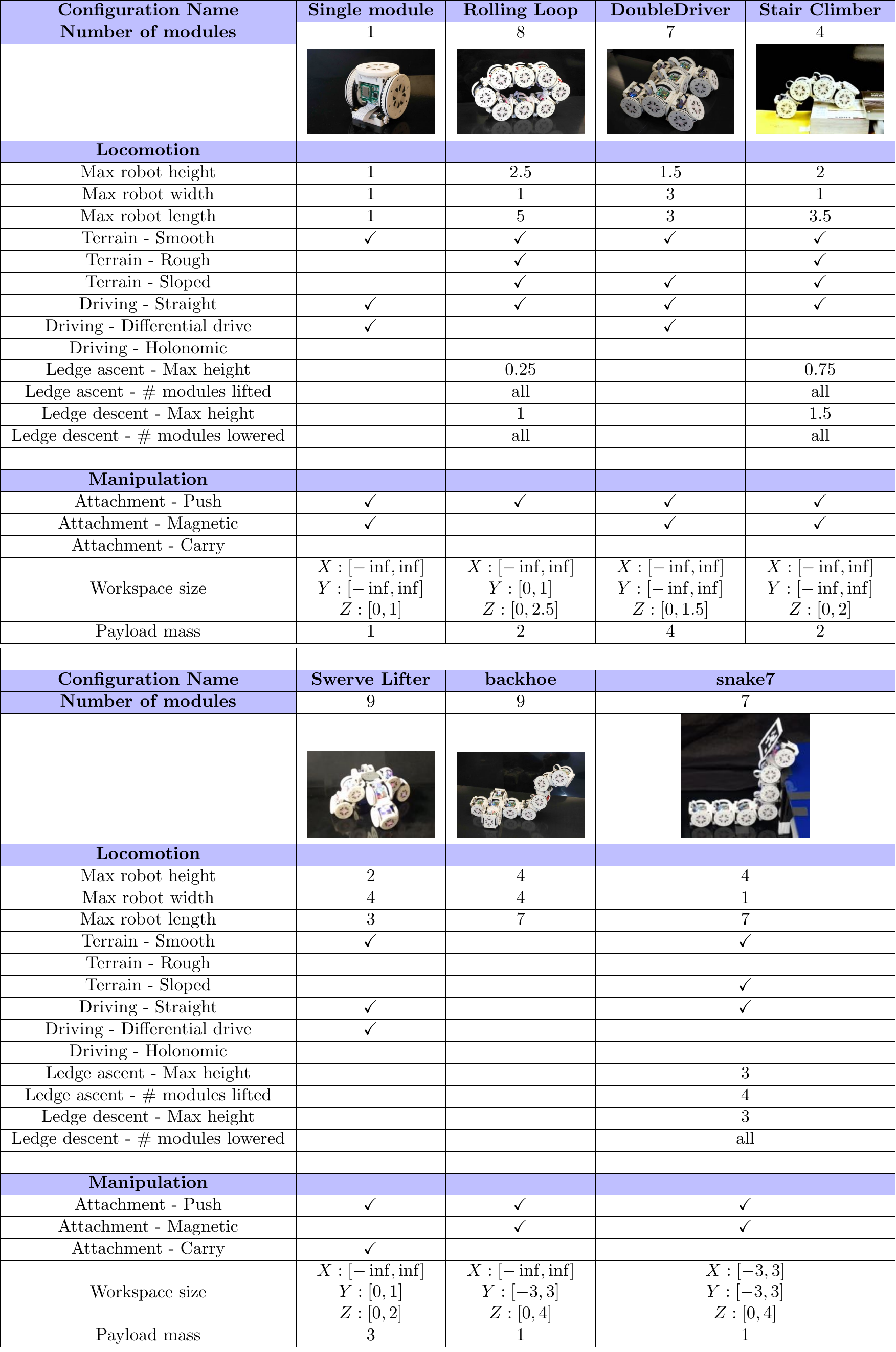}
        \caption{Matrix of designs and properties. Length and mass units are module-lengths and module-masses.}
        \label{tab:library-table}
\end{table*}
\subsubsection{Environmentally Adaptive Parametric Behaviors}
\label{sec:parametric}
In this section, we introduce an extension to the formalism we presented in \cite{jing2016end}.  As explained in Section~\ref{sec:background}, standard behaviors are defined as a series of joint angles or joint velocities for the modular robot cluster.  These are discrete, open-loop actions can be sequenced by the high-level mission planner to complete tasks.  Here, we present \emph{environmentally-adaptive parametric behaviors} (EAP behaviors) which provide additional functionality, allowing low-level behaviors to directly respond to sensed conditions in sophisticated ways. These behaviors are \textit{parametric}  because they take input arguments, called parameters, which allow them to produce a continuous range of motions. They are \textit{environmentally adaptive} because their parameters are intelligently assigned as a function of the state of the robot and environment.

\begin{definition}[Environmentally Adaptive\\Parametric Behavior]

We define an Environmentally-Adaptive Parametric Behavior as $B^{EAP}_C = (\{b_1, b_2, \ldots, b_n\},\mathbf{p},f)$, where $\{b_1, b_2, \ldots, b_n\}$ is a sequence of behavior states, $\mathbf{p} \in \mathbb{R}^m$ is a vector of parameters, and $f: \mathbb{R}^k \to \mathbb{R}^m$ is the controller function, where $\mathbb{R}^k$ is the space representing information about the robot and environment.
\end{definition}

Like standard behaviors, EAP behaviors consist of a sequence of behavior states $\{b_1, b_2, \ldots, b_n\}$.  However, some of the joint commands of these states are \emph{parametric}: instead of encoding fixed joint angles or velocities, they introduce a variable (called a \emph{parameter} of the behavior) that can be assigned their value whenever the behavior is called. Additionally, we associate with each EAP behavior a \emph{controller function} $f: \mathbb{R}^k \to \mathbb{R}^m$, which takes as input information about the robot and environment and produces as output the parameters of the behavior.  This function is a feedback controller which lets the behavior adapt to environment conditions. 

EAP behaviors expand the capabilities of our system.  For example, consider a single SMORES-EP module, which can drive on smooth terrain using its two wheels.  Using VSPARC, we can create a parametric \texttt{Drive} behavior that commands it to turn its wheels, assigning the wheel velocities to two parameters, e.g. $\mathbf{p} = \left\{ V_{\text{left}}, V_{\text{right}} \right\}$.  Using Python, we can now write a controller function for path following, taking as input the current location of the module and producing as output appropriate parameter values (wheel velocities) to drive the module along the path. In Section~\ref{hardware-demos}, we demonstrate how a similar \texttt{Drive} behavior and a path planner are used to direct a module to explore different regions on a tabletop.

As another example, consider the \texttt{Backhoe} configuration in Table~\ref{tab:library-table}. Using VSPARC, we can create a behavior that assigns parameters to the angles of all pan and tilt joints of the arm, providing access to the 7-DOF  forward kinematics of the robot.  For the controller function, we can write code that takes the position of an object as input and solves an inverse kinematics problem, providing output joint angles that cause the arm to touch an object.   

As the above examples imply, EAP behaviors have a two-step design process.  First, a parametric behavior is created using VSPARC, which we have extended to allow users to assign any joint value to a parameter rather than a fixed value.  This process is no more difficult than creating a non-parametric behavior. Next, a controller function is written, to provide the mapping from sensor data to parameter values. Controller functions can be quite sophisticated (examples include motion planners and feedback controllers) and are typically written in Python by an expert user.  However, if existing controller functions are available, novice users can re-use them to produce new EAP behaviors.  For example, the path-following controller developed for a single module could be re-used by a novice user to create a similar behavior for the \texttt{DoubleDriver} configuration (Table~\ref{tab:library-table}), which is also capable of differential drive.

\subsection{Reactive Controller Synthesis and Execution with the Library}

\begin{figure}
    \includegraphics[width=3.3in]{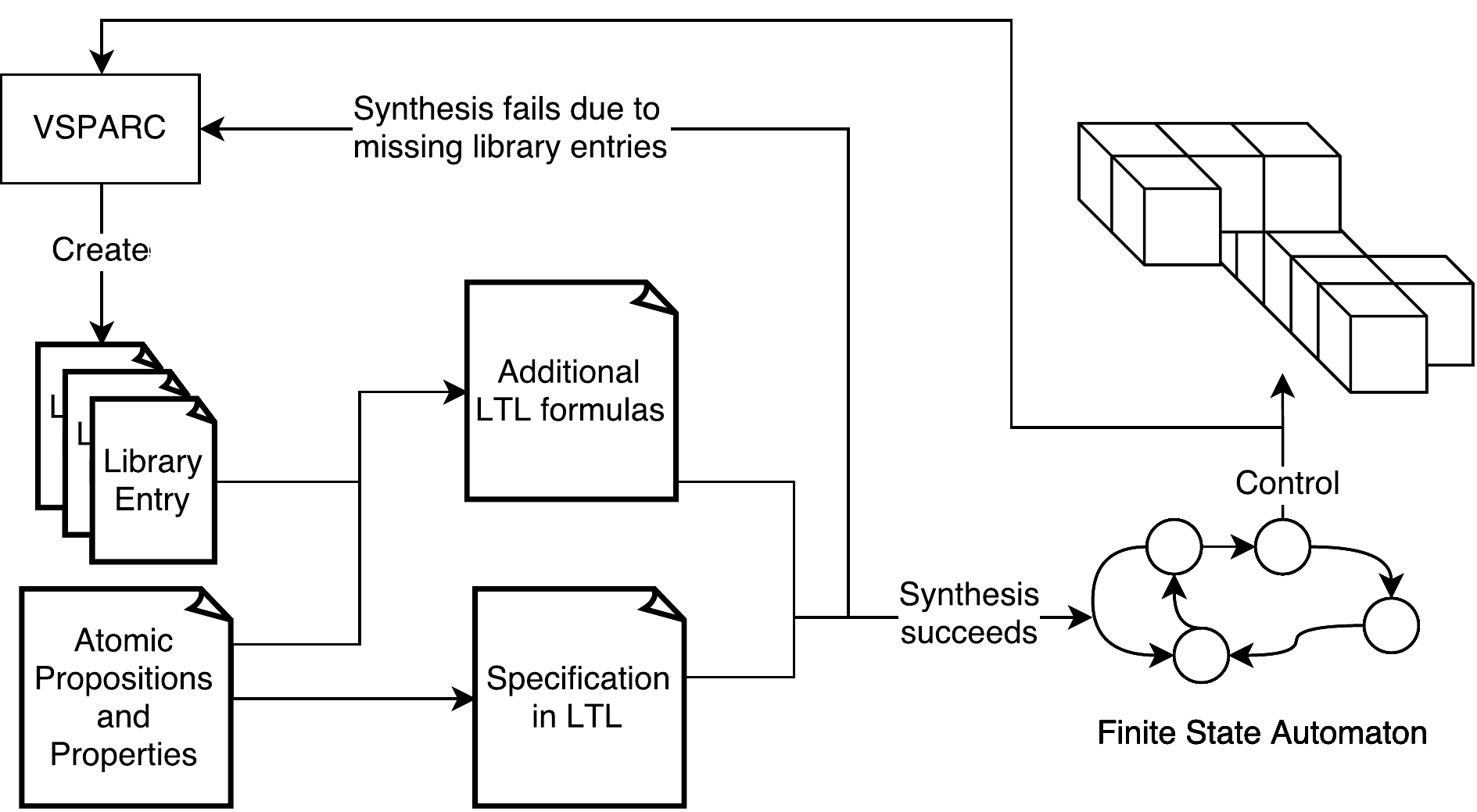}
    \caption{Controller synthesis and execution}
    \label{fig:abstraction2}
\end{figure}

In this section, we describe how our high-level mission planner synthesizes and executes controllers capable of accomplishing
tasks using configurations and behaviors from the design library. This process has three parts: (1) matching library entries with boolean variables, (2) generating additional LTL constraints imposed by mapping, and (3) executing the controller.
This framework is illustrated in Figure~\ref{fig:abstraction2}, and described in the following subsections.



%
\subsubsection{Matching library entries with boolean variables} \label{sec:matchprops}
As discussed in Section~\ref{sec:backgroundsynthesis}.
, users specify tasks using robot and environment variables that abstract robot actions and environmental conditions, as well as the mapping from these variables to low-level robot controllers.
Unlike conventional robots, modular robot systems can have multiple configurations and behaviors with similar capabilities.
Rather than providing a mapping  to specific behaviors, users label each variable in the task specification with sets of behavior and environment properties from the design library, to encode the desired functionality and constraints.
Our system searches the design library for a set of library entries that satisfy
the properties, and maps them to the corresponding boolean variable.
Consider an example robot task specification:
$$\texttt{if the robot senses {\bf Cup} then do {\bf Push}}.$$
The robot variable {\bf Push} might be described with:
\vspace{-0.4cm}
\begin{center}
\begin{gather*}
P = \{~\{ ( \text{Cup\_Mass},[1,3]) \}, \\
\{ (\text{Action},[\text{Drive}]), (\text{Speed},[1]) \}~\}
\end{gather*}
\end{center}
indicating that robot needs to be able to drive with speed of 1 with a cup that weights 1 to 3 module-weights.
With this specification, we can search through the robot design library to find a set of library entries $L_y=\{l_1, \dots, l_k\}$ that satisfies all properties in the set~$P$.

%


%

\subsubsection{Generating additional LTL formulas imposed by matching} \label{sec:generateLTL}
During the matching process,  additional necessary LTL constraints are automatically created among the
robot variables.
Consider a set of robot boolean variables $\mathcal{Y}$ used in a task specification.
We define a mapping relation $\lambda: \mathcal{Y} \rightarrow 2^{\mathcal{L}}$ that maps each variable $y\in \mathcal{Y}$ 
to a set of library entries $L_y$ that satisfies the user specified set of properties $P$ for $y$.
We say a library entry $l$ can {\it implement} a variable $y$ if $l \in \lambda(y)$.
For any $y\in \mathcal{Y}$, if $\lambda(y) = \emptyset$, we need to make sure variable $y$ is never \lt , because no library entry can implement $y$.
For example, users specify the robot action variable {\bf ClimbHigh} to be $P = \{~\{ ( \text{Height},[1,6]) \} \{ (\text{Action},[\text{Climb}])\}~\}$, and no matching behavior is found from the design library to implement the variable.
The system will generate addition LTL constraint to guarantee {\bf ClimbHigh} be \lf \; at all times.
Since there maybe multiple robot controllers that satisfy the given task specifications, the additional LTL constraint will force the robot to satisfy the task without ever perform the action {\bf ClimbHigh}, if possible.
If not, the additional LTL constraint will result in a failure to find a satisfying robot controller, in which case users need to modify the specification or design new robot behaviors.
For any $y,y'\in \mathcal{Y}$, if $\lambda(y) \cap \lambda(y') = \emptyset$, we need to make sure variable $y$ and $y'$ can never be \lt \ at the same time,
because there does not exist a library entry that can implement both $y$ and $y'$.
For example, additional constraints may be required to guarantee variables {\bf ClimbHigh} and {\bf Stop} are never \lt \; at the same time.
To encode the mutual exclusion between robot variables into the task specification, we specify them in the form of LTL formulas that are used together with the original task specification to generate robot controllers during synthesis.

%
%
%
%

\subsubsection{Controller Execution}\label{sec:controllerExe}
The synthesized finite-state automaton can be used to control simulated
or the physical robots.
If synthesis fails, possibly due to lack of
library entries that implement some robot variables,
LTLMoP will notify the user, who can then design suitable configurations and behaviors with
VSPARC.

A synthesized controller is executed by running behaviors based on the value of each robot action variable.
If a variable maps to a non-parametric behavior, the behavior is simply executed when the variable becomes \lt.
A behavior is stopped when the corresponding variable becomes \lf.

To execute an environmentally-adaptive parametric behavior, the values of all parametric joint commands are decided during execution by calling the controller function each time the behavior is executed.
For example, if the robot variable {\bf Explore} matches with the EAP \texttt{Drive} behavior of the \texttt{Single Module} configuration, the behavior will be executed whenever {\bf Explore} is \lt.
A path planner function computes values of parameters in \texttt{Drive} in order to control the robot as a two-wheel differential-drive car.

If two consecutive behaviors must be satisfied by two different configurations, reconfiguration is required. 
To reduce overall mission time, when multiple behaviors match with a robot boolean variable, we avoid unnecessary reconfiguration by biasing towards the behavior that requires no reconfiguration.

\section{Experimental Results}\label{sec:results}

We validate the capabilities of our system through experiments in simulation and hardware, illustrated in Figures~\ref{fig:simDemo}, \ref{fig:box-moving}, and \ref{fig:table-cleaning}, as well as the attached video\footnote{Video is also available online: \url{https://youtu.be/0rtXv4Z1E-o}}.
Faced with various task requirements, the system responds by synthesizing appropriate solutions.  The simulation experiments demonstrate how the high-level mission
planner can automatically synthesize and execute solutions to tasks using configurations and behaviors
from the library.
The hardware experiments validate that the system is capable of accomplishing complex physical tasks, such as carrying objects and climbing ledges.
\subsection{Simulated Task Scenarios}\label{sec:simulated-demos}
We present two simulated task scenarios.  A straightforward task is matched with a simple solution that uses one configuration, while a more complex task is addressed by reconfiguring between three different configurations, to leverage their wide-ranging capabilities.  
\subsubsection{Scenario 1}
In Scenario 1, our system must solve a multi-part task in the environment shown at the top of Figure~\ref{fig:simulator-scenarios}. The environment includes a button, a lightweight block, a gap in the ground, and a ramp, all in a straight line.  Pressing the button causes the block to drop to the ground, where it can be pushed into the gap, forming a bridge between the flat region and ramp.  When the task begins, the robot  is initially positioned in front of the button.  The objective is to reach a goal area at the top of the ramp.  The high-level action definitions for this task are provided in Table~\ref{tab:scenario1}.

After searching the library, the high-level mission planner discovers that the \texttt{rollingLoop} configuration has behaviors that satisfy the requirements of all three actions needed for this task (See Table~\ref{tab:library-table}).  To complete the task, the mission planner synthesizes a controller that commands the loop to press the button, push the block into the gap, and ascend the ramp, as shown in Figure~\ref{fig:simulator-scenarios}.

In response to this straightforward task, our system produces a simple solution.  As discussed in Section~\ref{sec:controllerExe}, the system attempts to minimize reconfiguration when completing a task, and so will opt to solve the entire task with a single configuration whenever possible.

\setlength\tabcolsep{5pt}
\begin{table}
\begin{center}
\begin{tabular}{| l | l |}
\hline
\textbf{Action Definition} & \textbf{Properties} \\
\hline
\texttt{pushButton:} & type = Manipulation\_Push \\
                     & height = $1.5$ \\
                     \hline
\texttt{pushBox:}    & type = Manipulation\_Push \\
                     & payload = $2$ \\
                     & distance\_x = $3$ \\
                     \hline
\texttt{climb:}      & type = Locomotion \\
                     & drive = Straight \\
                     & terrain = Sloped \\
\hline
\end{tabular}
\caption{High-level Action Definitions for Scenario 1}
\label{tab:scenario1}
\end{center}
\end{table}
\subsubsection{Scenario 2}
Like Scenario 1, Scenario 2 requires the robot to move from a starting position to a goal position.  However, several small changes have been made to the environment that makes the task more difficult.  The button has been moved to the side of the map, and floats at a height of 4 module-lengths above the ground.  The box is twice as heavy, weighing 4 module-weights rather than 2. The ramp has been replaced with stairs with a step height of $0.75$ module-lengths.  Table~\ref{tab:scenario2} provides the high-level action definitions for this scenario.

\begin{figure}
\begin{center}
    \includegraphics[width=\columnwidth]{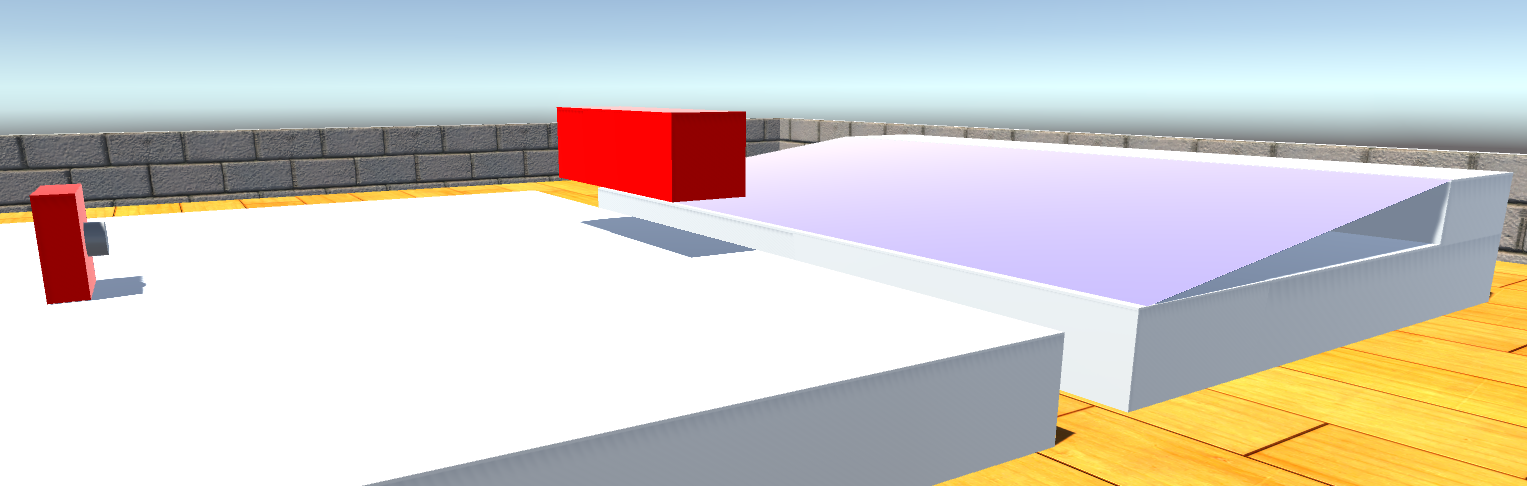}
    \includegraphics[width=\columnwidth]{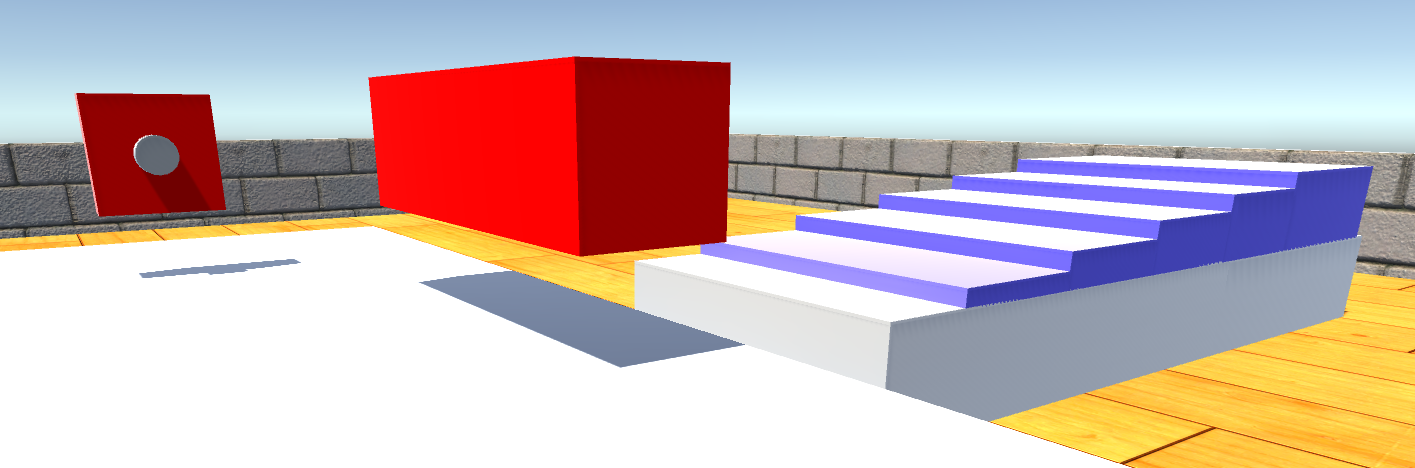}
    \caption{Environments for Scenarios 1 (top) and 2 (bottom) in the simulator.}
    \label{fig:simulator-scenarios}
\end{center}
\end{figure}

These changes make it impossible for the \texttt{rollingLoop} to complete the task - it can't reach the button, it's not strong enough to push the block, and it can't ascend steps more than $0.25$ module-lengths high. Instead, the high-level planner compiles a more complicated controller that uses behaviors from three different configurations in the library, shown in Figure~\ref{fig:simulator-scenarios}.  To push the button, the planner selects the \texttt{backhoe}, because it is the only configuration with a large enough vertical workspace.  To push the block into the gap, the robot reconfigures into the \texttt{doubleDriver}, which is capable of driving, turning, and pushing objects as heavy as 5 module-weights.  To climb the stairs, the robot reconfigures into the \texttt{stairClimber}, which can easily ascend $0.75$ module-length steps.

This scenario demonstrates how our system leverages the flexibility of modular robots.  This challenging task requires the diverse capabilities provided by all three configurations, and could not be accomplished by any one of them alone.  Note that for the purposes of this work we do not provide strategies to autonomously perform self-reconfiguration. Instead, we assume that the robot can self-reconfigure between any two configurations as long as the initial configuration has an equal or greater number of modules than the final configuration.  This does not fundamentally limit the power of our system: techniques for autonomous self-reconfiguration with SMORES-EP have been recently developed, and could easily be incorporated \cite{daudelin2017}.
%
\begin{table}
\begin{center}
\begin{tabular}{| l | l | }
\hline
\textbf{Action Definition} & \textbf{Properties} \\
\hline
\texttt{pushButton:} & type = Manipulation\_Push \\
                     & height = $4$ \\
                     \hline
\texttt{pushBox:}    & type = Manipulation\_Push \\
                     & payload = $4$ \\
                     & distance\_x = $3$ \\
                     \hline
\texttt{climb:}      & type = Locomotion \\
                     & drive = Straight \\
                     & ledge height = $0.75$\\
                     \hline
\end{tabular}
\caption{High-level Action Definitions for Scenario 2}
\label{tab:scenario2}
\end{center}
\end{table}
\setlength\tabcolsep{0pt}
%
%
%
\subsection{Hardware Experiments}\label{hardware-demos}
\begin{figure}[h]   
    \begin{center}
    \includegraphics[width=\columnwidth]{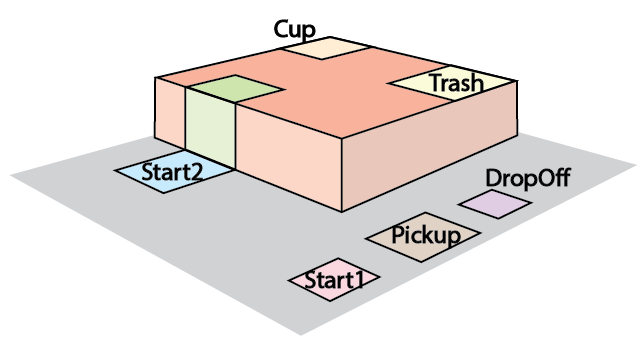}
    \end{center}
    \caption{Map of the hardware demo}
    \label{fig:map}
\end{figure}

Our hardware experiments demonstrate that our system can accomplish a complex physical task using physical SMORES-EP robot modules. The robot is required to clean the top of a table, operating in the environment shown in Figure~\ref{fig:map}. To do so, the robot must first move a waste bin from its initial location (labelled ``Pickup'') to a target location next to the table (labelled ``Dropoff'').  Then, the robot must climb to the top of the table and explore the surface.  Whenever it encounters an object, it must react appropriately: if it is garbage, it should push it off the table and into the waste bin, and if not, it should notify a human to remove it. 

This experiment showcases the seamless translation of behaviors from the VSPARC simulator to hardware, and the ability to use the LTLMoP high-level planner to create mission plans that can be directly executed by the modules.  AprilTags tracked by an overhead
camera provide information about the position of modules and objects in the environment,
serving as sensory feedback for the high-level planner.

This experiment also demonstrates how the design library is continually expanded as users develop designs to address new tasks.  While the library encompasses a wide range of functionality, it is by no means complete: when a high-level specification was first created for this experiment, the mission planner reported that it could not be satisfied using existing elements in the library.  Consequently, two new configurations (the \texttt{swerveLifter} and \texttt{snake7} configurations) were created, and low-level behaviors were iteratively developed to fulfill the needs of each component of the task.  Once these configurations and behaviors were made available in the library, the high-level planner was able to successfully synthesize and execute controllers to accomplish the tasks.
\subsubsection{Moving the Waste Bin}\label{moving-the-box}
The robot begins its task in region \texttt{Start1}, and must move the waste bin from \texttt{Pickup} to \texttt{Dropoff}, a distance of 10 module lengths.  Once the waste bin is in place beside the table, the robot must travel to the edge of the table (\texttt{Start2}), where it can begin the next phase of the task (exploring the tabletop).

The waste bin is a box supported by four legs, making it impossible for any design less than two module-heights tall to push it.  This constraint rules out most car-like configurations in the library.  The 10-module distance over which the bin must be transported imposes a workspace requirement that rules out all stationary manipulators. Consequently, the \texttt{swerveLifter} configurations was designed to meet all the criteria.  The \texttt{swerveLifter} uses four SMORES-EP modules as powered caster wheels, allowing omnidirectional movement (sometimes called \emph{swerve drive}). It can also raise and lower, enabling it to lift and carry objects by driving underneath them.

The high-level description of this phase of the task is shown in Specification~\ref{spec:wastebin}, and Figure~\ref{fig:box-moving} shows how the robot completes it.  The task is reactive: the robot waits until it senses the waste bin before beginning the \textbf{pickup} action (Line 3 of Specification~\ref{spec:wastebin}). Once the waste bin appears (i.e. the AprilTag marking it comes the camera view), the robot lowers itself, drives beneath the waste bin, and carries it to the \texttt{Dropoff} region.  It then moves back out from beneath the waste bin, and executes a series of omnidirectional driving behaviors to travel to the edge of the table.
%
\begin{spec}[h!]
\caption{Moving the Wastebin}
\label{spec:wastebin}
\vspace{-0.1cm}
\small\setlength{\jot}{0pt}
\begin{fleqn}[3pt]
\leqnomode
\begin{subequations}
\renewcommand{\theequation}{\arabic{equation}} 
\makeatletter
\renewcommand\tagform@[1]{\maketag@@@{\ignorespaces#1\unskip\@@italiccorr}}
\makeatother
\hskip-10cm
\begin{alignat}{2}
&\text{{\bf 1.} \textbf{carry} is set on \textbf{pickup} and reset on false} && \notag
\\
&\text{{\bf 2.} \textbf{dropped} is set on \textbf{drop} and reset on false} &&
\notag
\\
&\text{{\bf 3.} do \textbf{pickup} if and only if you were sensing \textbf{wasteBin}}&& \notag
\\
&\hspace{1cm}\text{{and you are not activating \textbf{carry}}}&& \notag
\\
&\text{{\bf 4.} do \textbf{goToTable} if and only if you are activating \textbf{dropped} }&& \notag
\\
&\text{{\bf 5.} do \textbf{drop} if and only if you were activating \textbf{carry}}&& \notag
\\
&\hspace{1cm}\text{{and you are not activating \textbf{dropped}}}&& \notag
\end{alignat}
\end{subequations}
\end{fleqn}
\vspace{-0.4cm}
\end{spec}
\subsubsection{Table Exploration}\label{table-exploration}
With the waste bin in place, the robot begins the second phase of the task: cleaning the top of the table.  The robot needs to climb to the tabletop, explore, and react to what it finds.  The \texttt{snake7} configuration was designed to be capable to do this. As shown in Table~\ref{tab:library-table}, the \texttt{snake7} configuration can use its \texttt{climbup} and \texttt{climbdown} behaviors to ascend and descend ledges up to 3 module-heights tall.  However, it is unable to lift its entire body up to the tabletop, and even if it could, it would be too large to effectively explore.  Instead, the robot reconfigures, detaching the front module of the snake to act as a \texttt{module1} configuration that can use its EAP behavior \texttt{differentialDrive} to explore the tabletop, and its  \texttt{spin}, and \texttt{push} behaviors to clean.

Specification~\ref{spec:tabletop} provides the high-level task description, and Figure~\ref{fig:table-cleaning} shows the robot completing the task.  The robot begins in the \texttt{snake7} configuration, positioned at the edge of the table in the \textbf{ground} region.
An AprilTag is fixed to the front module of the snake, allowing the mission planner to determine its location at all times. Sensing that it is in the \textbf{ground} region, the \texttt{snake7} executes \textbf{climbup} (line 8 of Specification~\ref{spec:tabletop}). After climbing, the mission planner senses that the head of the snake has reached the \textbf{dock} region at the edge of the tabletop, and executes the \textbf{undock} behavior to detach the head module from the snake (line 6), allowing it to operate on its own as a \texttt{module1}.

The module then uses \texttt{differentialDrive} to visit two regions of interest on the tabletop (\textbf{loc1} and \textbf{loc2}). \texttt{differentialDrive} is an EAP behavior that allows the robot to explore its environment in a continuous fashion.  The driving behavior and its parameters are the same as the driving behavior presented as an example in Section~\ref{sec:parametric}: two parameters specify the left and right wheel velocities.  The controller function is a potential field path planner that maps the robot's current position (sensed via AprilTag) to a desired linear and angular velocity, which are converted to wheel velocities.

When it reaches \textbf{loc1}, the robot senses a coffee mug (marked with an AprilTag), and responds by executing a \textbf{spin} behavior to notify a nearby human that it should be removed (line 1).  When it reaches \textbf{loc2}, it senses a piece of trash, and it correctly responds by performing a \textbf{push} to move it off the table and into the waste bin.
Having fully explored the table, the module returns to the dock point and re-attaches to the body of the snake (line 5).  The snake then executes \textbf{climbdown} to descend back to the floor, completing its mission.
%
\begin{spec}[h!]
\caption{Cleaning the Tabletop}
\label{spec:tabletop}
\vspace{-0.1cm}
\small\setlength{\jot}{0pt}
\begin{fleqn}[3pt]
\leqnomode
\begin{subequations}
\renewcommand{\theequation}{\arabic{equation}} 
\makeatletter
\renewcommand\tagform@[1]{\maketag@@@{\ignorespaces#1\unskip\@@italiccorr}}
\makeatother
\hskip-10cm
\begin{alignat}{2}
&\text{{\bf 1.} if you are sensing \textbf{mug} then do \textbf{spin}} && \notag
\\
&\text{{\bf 2.} if you are sensing \textbf{trash} then do \textbf{push}} &&
\notag
\\
&\text{{\bf 3.} \textbf{loc1visited} is set on \textbf{loc1} and reset on false}&& \notag
\\
&\text{{\bf 4.} \textbf{loc2visited} is set on \textbf{loc2} and reset on false }&& \notag
\\
&\text{{\bf 5.} do \textbf{docking} if and only if you were in \textbf{dock} and you}&& \notag
\\
&\hspace{1cm}\text{{are activating
  (\textbf{loc1visited} and \textbf{loc2visited})}}&& \notag
\\
&\text{{\bf 6.} do \textbf{undock} if and only if you were in \textbf{dock} and you}&& \notag
\\
&\hspace{1cm}\text{{are not activating
  (\textbf{loc1visited} or \textbf{loc2visited})
}}&& \notag
\\
&\text{{\bf 7.} do \textbf{climbdown} if and only if you were in \textbf{dock} and}&& \notag
\\
&\hspace{1cm}\text{{you activated
  (\textbf{loc1visited} and \textbf{loc2visited})
}}&& \notag
\\
&\text{{\bf 8.} do \textbf{climbup} if and only if you were in \textbf{ground} and}&& \notag
\\
&\hspace{1cm}\text{{you are not
  activating (\textbf{loc1visited} or \textbf{loc2visited})
}}&& \notag
\\
&\text{{\bf 9.} infinitely often do \textbf{docking}}&& \notag
\end{alignat}
\end{subequations}
\end{fleqn}
\vspace{-0.4cm}
\end{spec}

\subsubsection{Challenges}

In general, the hardware experiment was successful, with the high-level planner successfully executing library behaviors to complete this task.  While running the experiment, several notable challenges were encountered.  During the first phase (moving the waste bin), achieving accurate steering with the \texttt{swerveLifter} proved difficult.  The \texttt{swerveLifter} steers by aligning four caster wheels in the same direction, a process that is sensitive to encoder calibration errors across modules.  Recently, more sophisticated calibration procedures for the SMORES-EP encoders have been developed, and encoder performance has been improved \cite{Tosun2017paintpots}.

During the second phase of the experiment (exploring the tabletop), careful initial positioning was required for the open-loop \texttt{climbUp} behavior to succeed - in several trials, the snake was started too close to the ledge, causing it to collide with the corner of the table and break. This problem could be alleviated by developing an EAP behavior allowing the robot to autonomously drive to the appropriate distance before beginning to climb.

In both phases of the experiment, limited magnetic connector strength between modules presented a significant challenge.  The \texttt{swerveLifter} configuration had to be constructed with a passive cube in its center in order to perform its raising and lowering behaviors without breaking. During descent from the table, bending forces experienced at the center of the \texttt{snake7} configuration would sometimes cause connections between modules to break.

The limited strength of the magnetic connectors can be viewed as a trade-off for ease of reconfiguration.  Connection and disconnection between the head and body of the snake takes very little time, and the forgiving area-of-acceptance of the connector \cite{Tosun2016} makes it possible to dock the head of the snake to the body even though the exact position of the body is not known (only the head module had an AprilTag).  Autonomous docking succeeded about 25\% of the time. This performance could be improved by applying more recently developed techniques for autonomous self-reconfiguration with SMORES-EP.  recently, improved docking strategies for SMORES-EP have been developed that succeed about 90\% of the time. \cite{daudelin2017}.
%
\newcommand{\figwidth}{\textwidth/4}
\newcommand{\textsize}{}
\setlength\tabcolsep{0pt}
\begin{figure*}
\begin{center}
\begin{tabular}{c|c|c|}
\hline
  \includegraphics[width=1.1\figwidth]{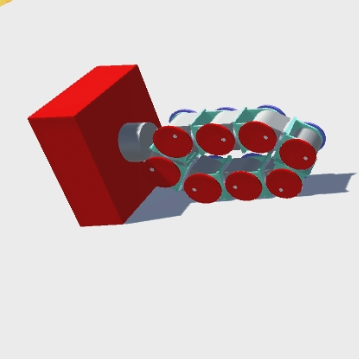} &
  \includegraphics[width=1.1\figwidth]{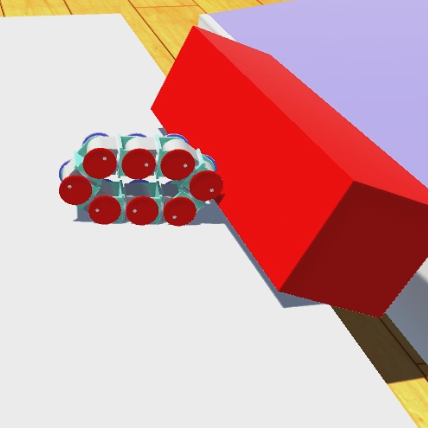} &
  \includegraphics[width=1.1\figwidth]{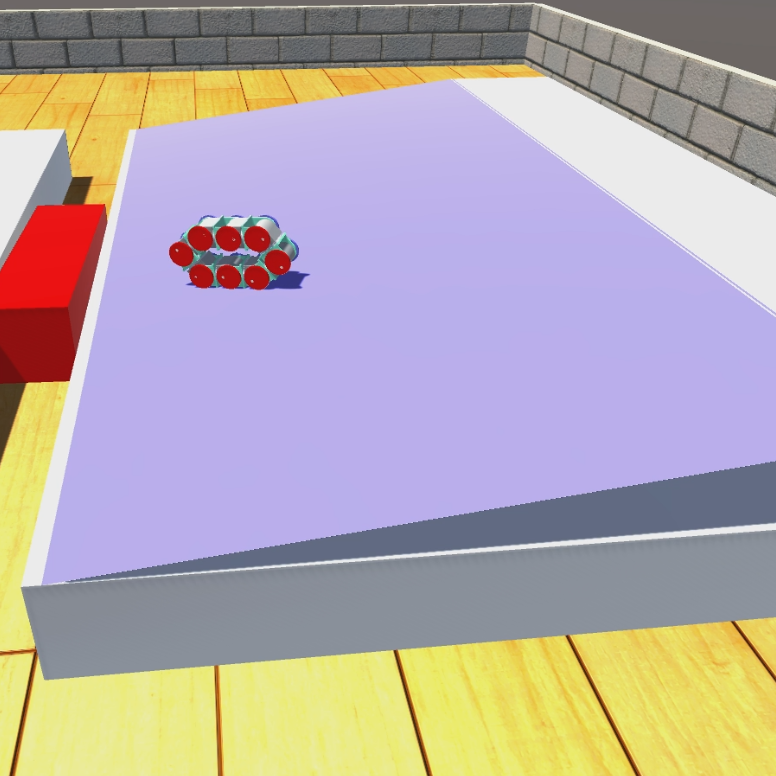}  \\
  \hline
  \textsize\texttt{rollingLoop.backward} &
  \textsize\texttt{rollingLoop.forward}  &
  \textsize\texttt{rollingLoop.forward}  \\
  \hline
  \includegraphics[width=1.1\figwidth]{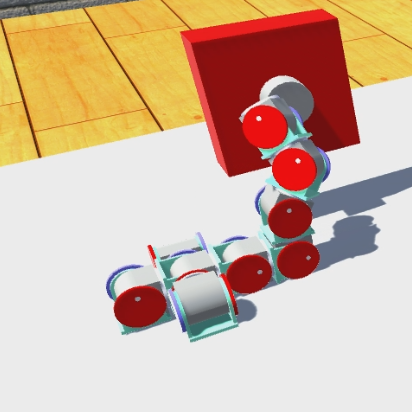} &
  \includegraphics[width=1.1\figwidth]{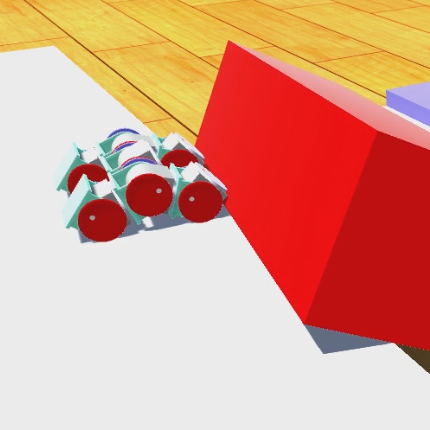} &
  \includegraphics[width=1.1\figwidth]{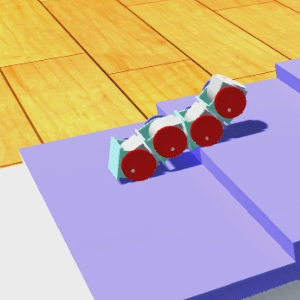}  \\
\hline

  \textsize\texttt{backhoe.pressButton}  &
  \textsize\texttt{doubleDriver.turnAndDrive}  &
  \textsize\texttt{stairClimber.climb}  \\
  \hline
\end{tabular}
\end{center}
\caption{Simulated Demo}
\label{fig:simDemo}
\end{figure*}
%
%
%
\begin{figure*}
\begin{center}
\begin{tabular}{c|c|c|c}
\hline
  \includegraphics[width=\figwidth]{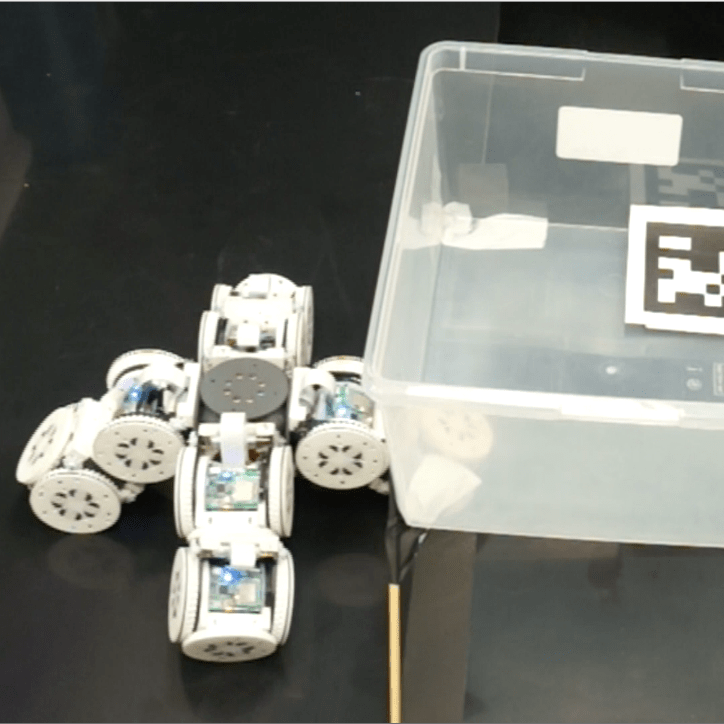}&
  \includegraphics[width=\figwidth]{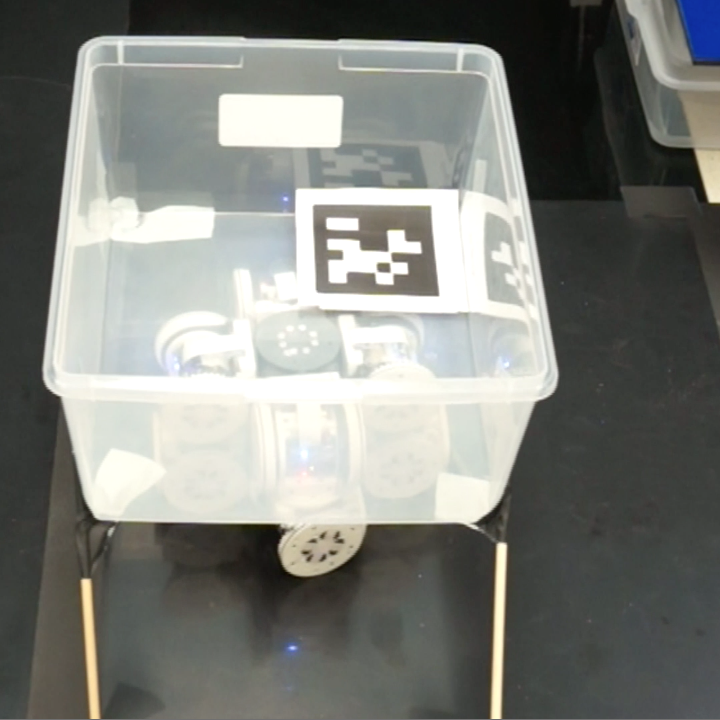} &
  \includegraphics[width=\figwidth]{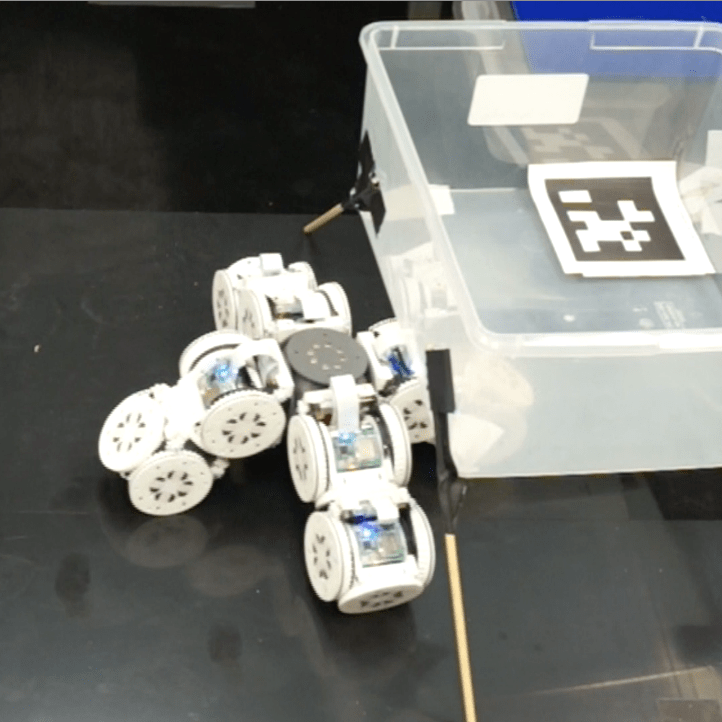} &
  \includegraphics[width=\figwidth]{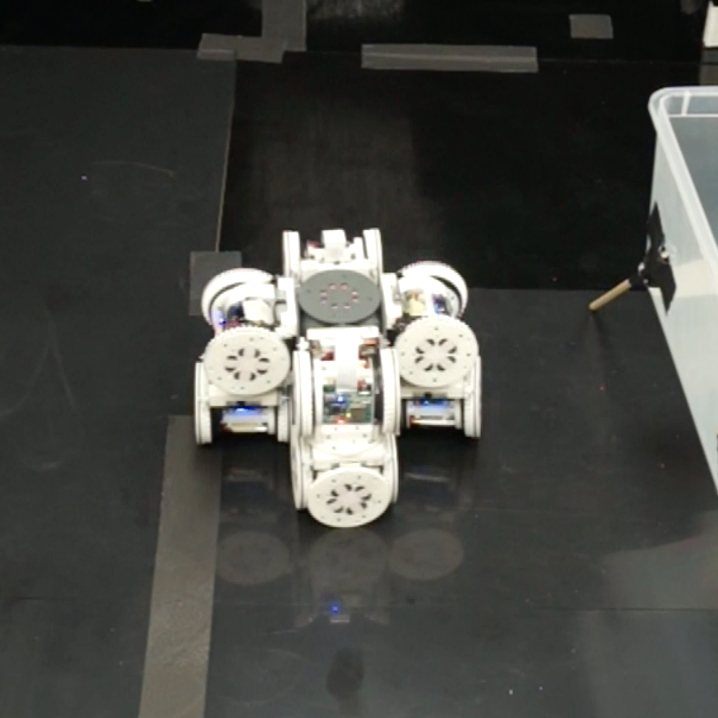} \\
\hline
  \includegraphics[width=\figwidth]{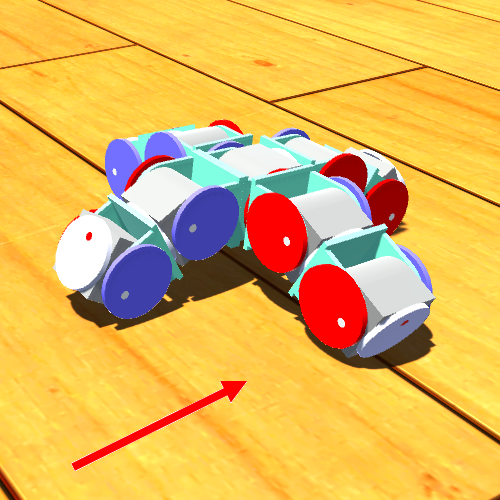} &
  \includegraphics[width=\figwidth]{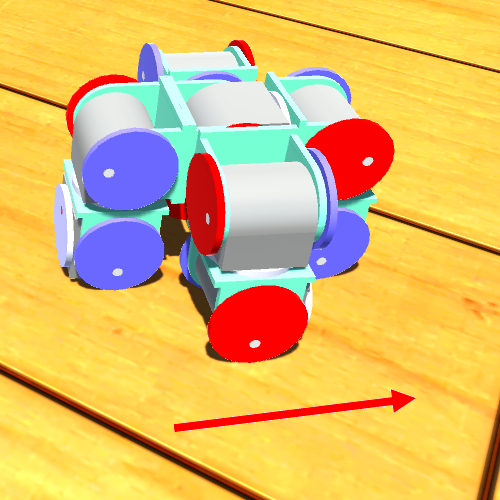} &
  \includegraphics[width=\figwidth]{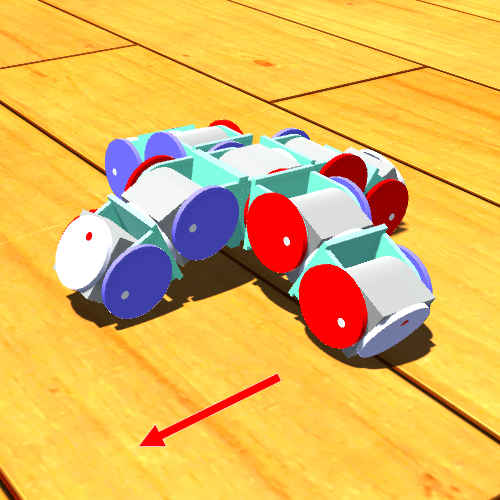} &
  \includegraphics[width=\figwidth]{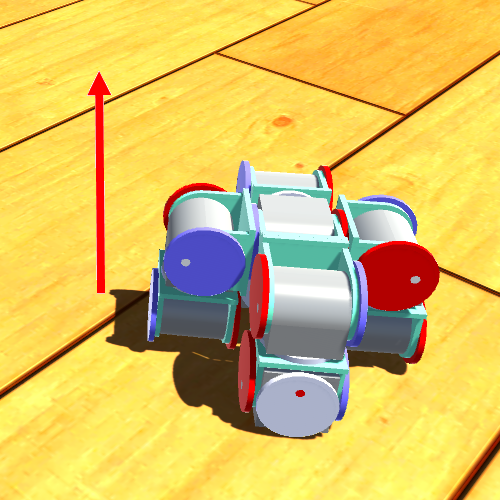} \\
\hline
  \textsize\texttt{swerveLifter.goUnder} &
  \textsize\texttt{swerveLifter.carry}  &
  \textsize\texttt{swerveLifter.dropOff}  &
  \textsize\texttt{swerveLifter.driveUp}\\
  \hline
\end{tabular}
\end{center}
\caption{Moving the Waste Bin}
\label{fig:box-moving}
\end{figure*}
%
%
\begin{figure*}
\begin{center}
\begin{tabular}{|c|c|c|c|}
\hline
  \includegraphics[width=\figwidth]{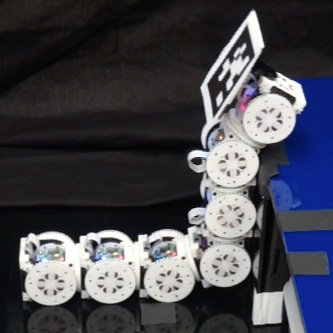}&
  \includegraphics[width=\figwidth]{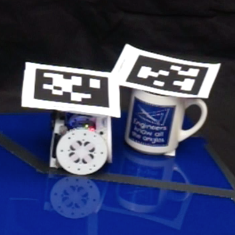} &
  \includegraphics[width=\figwidth]{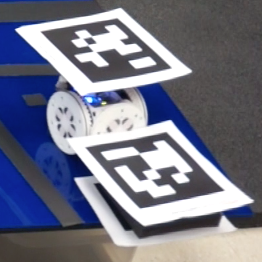} &
  \includegraphics[width=\figwidth]{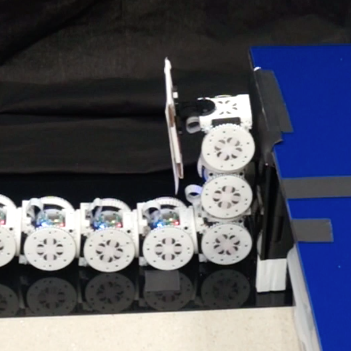} \\
\hline
  \includegraphics[width=\figwidth]{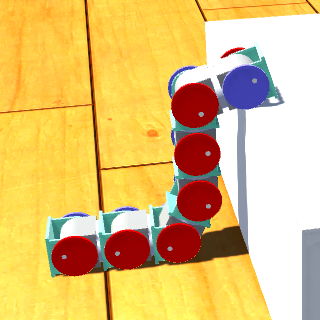} &
  \includegraphics[width=\figwidth]{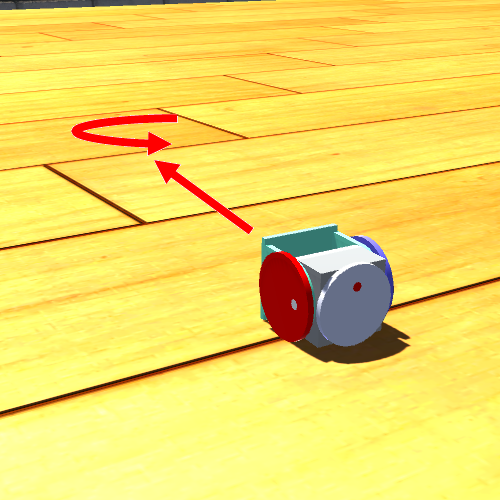} &
  \includegraphics[width=\figwidth]{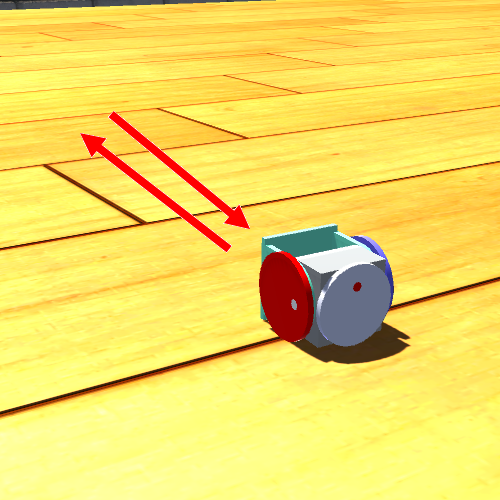} &
  \includegraphics[width=\figwidth]{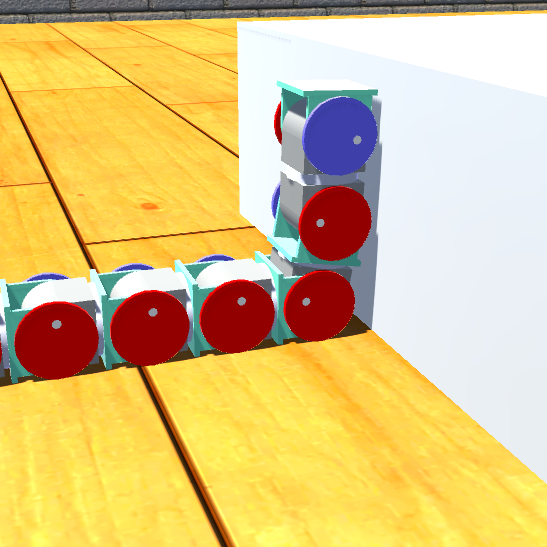} \\
\hline
  \textsize\texttt{snake7.climb} &
  \textsize\texttt{module.spin}  &
  \textsize\texttt{module.push}  &
  \textsize\texttt{snake7.descend}
\end{tabular}
\end{center}
\caption{Cleaning the Table}
\label{fig:table-cleaning}
\end{figure*}
%
%
%
\section{Discussion and Future Work} \label{sec:discussion}

\subsection{Simulator-to-hardware translation}
Translation of behaviors from VSPARC to the hardware was largely successful, and the ability to prototype designs and behaviors in a simulator resulted in significant time savings over prototyping in hardware.  Disparities between performance in the simulator and hardware tended to arise from real-world phenomena the simulator did not model accurately.  For example, variability in magnetic connector strength  (which differs from module to module \cite{Tosun2016}) sometimes resulted in connections breaking unexpectedly, and encoder calibration errors could cause behaviors requiring very precise position control to perform poorly.

Utilizing EAP behaviors can improve the robustness of behaviors designed with VSPARC.
Users can create behaviors whose joint commands depend on the robot encoder reading on the fly.
Moreover, incorporation of on-board sensing will allow our system to operate autonomously in unknown environments. At the time of writing, a ``brain module'' has been developed that allows SMORES-EP clusters to carry an RGB-D camera and computer unit \cite{daudelin2017}.
Using sensing information to decide joint values for EAP behaviors creates robust closed-loop behaviors.
In the future, VSPARC could be expanded to include simulated sensing capabilities, making it easier to develop closed-loop EAP behaviors.

\subsection{Library Creation: Lessons Learned}
Early on in the development of this system, we intended to populate our design library through crowdsourcing, using a system such as Amazon Mechanical Turk where a large number of online users could create configurations and behaviors using VSPARC.  We quickly realized that this strategy would not produce high-quality designs: developing sophisticated designs and behaviors in the simulator requires skill and experience.  Holding hackathons with undergraduate engineers proved to be a much more effective strategy, because participants would become significantly more adept at creating designs and behaviors through hours of practice.  Newcomers would typically spend about an hour creating a useful behavior, where well-practiced users would spend about twenty minutes.
Introducing EAP behaviors to VSPARC greatly expanded the design space and thus the capabilities of the designed robots.
A well-designed EAP behaviors can handle a wider range of environments, such as stairs with different heights, which would normally require users to create multiple static behaviors.
EAP behaviors can also be combined with environment perception to create complex behaviors such as obstacle avoidance and target following that are very hard to achieve with static behaviors. 

Interestingly, users spent significantly more time creating behaviors than configurations.  Most users required only a few minutes to build a new configuration and conceive of the fundamental motions they wanted it to perform.  The majority of the design time was spent coding joint trajectories to achieve the desired motion while maintaining balance and avoiding connector strength overload.
With EAP behaviors, users can focus the design on high-level without explicitly coding joint angles, which can reduce designing time.
Evolutionary techniques will also be explored to generate behaviors automatically.

An existing algorithm for modular robot design embedding detection could also be used to automatically generate behaviors for new configurations \cite{mantzouratos2015embeddability}.  This algorithm can automatically detect when one a subset of the joints of one configuration can be used to replicate the kinematics of another (a condition known as \emph{embedding}), and generates a mapping that can be used to transfer behaviors originally developed for one configuration to any other configuration that embeds it.  This could also allow behaviors developed for SMORES-EP to be ported to other modular robot systems, or vice-versa.  

\subsection{Composing Library Elements to Complete Missions}
Environment and behavior properties provide an expressive way for the user to
specify the requirements of a task. However, the fact that a behavior is
labeled with a specific property does not guarantee it will perform as
intended in all circumstances. Adapting behaviors to environments
different from the one in which they were designed can cause them to fail, as
evidenced by the problems in establishing proper initial robot position for the
\texttt{climbUp} behavior in the table cleaning scenario.  Development of more closed-loop parametric behaviors will help address this issue.

Methods for automatically analyzing tasks and environments are actively being researched \cite{sung2016robobarista}.  Determining optimal sets of environment factors and integrating methods for automatic task analysis and would be an interesting avenue for future work. 

It's worth noting that some behaviors are much more tolerant to varying environments than others.  In our hardware experiments with the \texttt{stairClimber} configuration, we found that a single open-loop gait was able to climb steps of several varying sizes with no problems.  Establishing confidence bounds on behavior success as a function of environment parameters and including this information in the library is future work.

In this work, we assume that reconfiguration is possible between any two configurations as long as the initial configuration has at least as many modules as the final configuration.  In practice, autonomous self-reconfiguration often requires complicated behaviors that have implications for high-level planning: for example, if the robot is holding an object, that may affect its ability to reconfigure.  Autonomous self-reconfiguration with SMORES-EP is addressed in another paper by the authors~\cite{daudelin2017}.  Other work on autonomous self-reconfiguration includes \cite{Yim2007},\cite{Rubenstein2004},\cite{Murata2006}.

Our architecture would rely on a library with a large number of behaviors and attributes in order to encode robot capabilities and environment properties for real-world experiments.
When searching the library for behaviors, our method scales linearly with respect to the size of the library.

%
%
\section{Conclusion} \label{sec:conclusion}
We presented a system for addressing high-level tasks with modular self-reconfigurable robots. We demonstrated how our physics-based simulator allows SMORES-EP configurations and behaviors to be easily created and stored in the design library, and how our framework for labeling each entry in the library with descriptive properties allows them to be organized by functionality.  Integration with a high-level mission planner allowed users to provide high-level task specifications, which were used to synthesize reactive controllers that use configurations and behaviors from the library. The capabilities of our system are validated through experiments in simulation and with physical modular robots.  Building beyond our earlier work, we also expanded the system by introducing environmentally-adaptive parametric behaviors, which allowed sophisticated motion planners and feedback controllers to be used within our framework.

This system is among the first to address complex, reactive, high-level tasks with modular self-reconfigurable robots. By providing this framework and demonstrating its success in the lab, we hope to lay the foundation for future modular robot systems to address tasks in the real world.
%
\section*{Acknowledgments}
This work was funded by NSF grant numbers CNS-1329620 and CNS-1329692.



\bibliographystyle{spmpsci}
\bibliography{references}

\end{document}